\documentclass[runningheads]{llncs}

\usepackage{eccv}

\usepackage{eccvabbrv}

\usepackage{graphicx}
\usepackage{booktabs}
\usepackage[accsupp]{axessibility}  
\usepackage{comment}

\usepackage{hyperref}

\usepackage{orcidlink}

\begin{document}

\title{Degradation-based augmented training for robust individual animal re-identification} 

\titlerunning{Degradation-based augmented training for animal re-ID}

\author{Thanos Polychronou\inst{1}\orcidlink{0000-0001-8225-5303} \and
Lukáš Adam\inst{2}\orcidlink{0000-0001-8748-4308} \and
Viktor Penchev\inst{1}\and\\
Kostas Papafitsoros\inst{1}\orcidlink{0000-0001-9691-4576}
}

\authorrunning{T.~Polychronou et al.}

\institute{
School of Mathematical Sciences, Queen Mary University of London, UK 
\and
University of West Bohemia in Pilsen, FEE, RICE, Pilsen, Czechia \\[0.5em]
\email{\{a.polychronou, v.penchev, k.papafitsoros\}@qmul.ac.uk},
\email{lukas.adam.cr@gmail.com}}

\maketitle

\begin{abstract}
Wildlife re-identification aims to recognise individual animals by matching query images to a database of previously identified individuals, based on their fine-scale unique morphological characteristics. Current state-of-the-art models for multispecies re-identification are based on deep metric learning representing individual identities by feature vectors in an embedding space, the similarity of which forms the basis for a fast automated identity retrieval. Yet very often, the discriminative information of individual wild animals gets significantly reduced due to the presence of several degradation factors in images, leading to reduced retrieval performance and limiting the downstream ecological studies. Here, starting by showing that the extent of this performance reduction greatly varies depending on the animal species (18 wild animal datasets), we introduce an augmented training framework for deep feature extractors, where we apply artificial but diverse degradations in images in the training set. We show that applying this augmented training only to a subset of individuals, leads to an overall increased re-identification performance, under the same type of degradations, even for individuals not seen during training. The introduction of diverse degradations during training leads to a gain of up to 8.5\% Rank-1 accuracy to a dataset of \emph{real-world degraded} animal images, selected using human re-ID expert annotations provided here for the first time. Finally, to showcase the advantage of capturing  domain-specific degradations, we employ a style-transfer GAN to generate synthetic yet realistic low-quality images that closely resemble the real-world samples in the annotated dataset. Using these images for data augmentation yields further improvements in re-identification performance compared to the previous handcrafted augmentation framework on the same dataset. Our work is the first to systematically study image degradation in wildlife re-identification, while introducing all the necessary benchmarks, publicly available code\footnote{Code is available at https://github.com/apolychronou/dba-reid} and data, enabling further research on this topic.

\end{abstract}

\section{Introduction}
\label{sec:intro}
\noindent
\textbf{Wildlife re-ID and image degradation}\\[0.5em]
Wild animal re-identification (re-ID) denotes the task of identifying individual animals from images taken across different encounters in the natural environment. It explores fine-grained discriminative information stemming from the animals' unique external morphological characteristics, e.g. spots, stripes, scales and other patterns. Wildlife re-ID is an indispensable tool for ecological research for various species, informing several aspects of it, e.g.\ population abundance, survival rates, behaviour, movement and anthropogenic threats \cite{parham2017animal, konrad2018kinship, Schofield_2020, papafitsoros_social_2021}. In practice, the goal is to match newly obtained images - the \emph{query images} - to a catalogue of images of previously identified individuals (identity retrieval).
During the last years, there has been significant progress from the fields of computer vision and machine learning on automatising and accelerating the procedure \cite{tuia2022perspectives, vidal2021perspectives, jiao2023toward}, see also \cite{cermak_wildlifedatasets_2024} for a detailed method categorisation. The current state-of-the-art on animal re-ID is evolving around \emph{deep feature-based approaches} (deep metric learning \cite{lu_deep_2017}) which learn discriminative representations from large-scale datasets using deep neural networks \cite{cermak_wildlifedatasets_2024, haurum_re-identification_2020, deb_face_2018,  otarashvili_multispecies_2024, hou2025openanimals}.\\ 
\begin{figure}[t]
    \centering
    \includegraphics[width=0.99\textwidth]{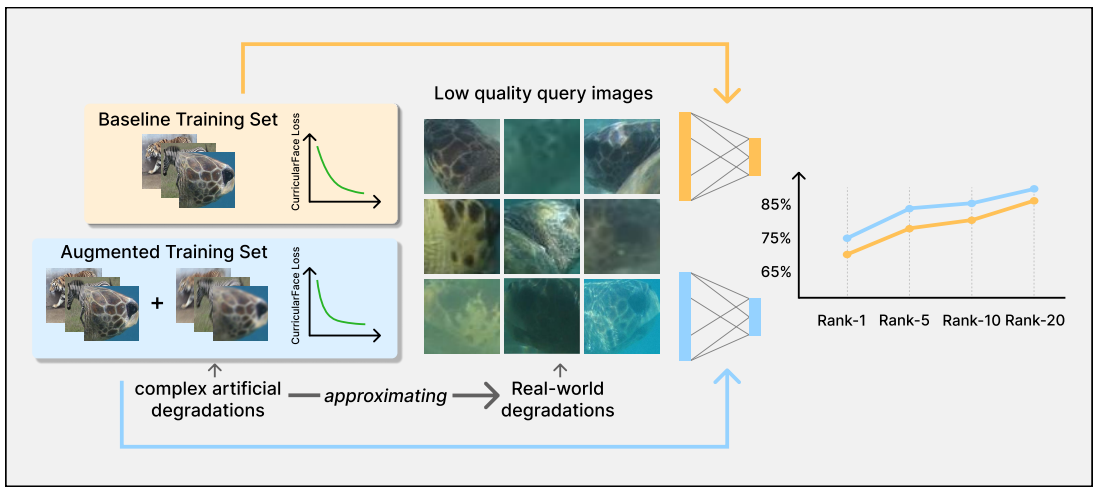} 
    \caption{Augmenting the training set of deep feature-based re-ID models, with artificial but complex degradations, which can be  model-based or data-driven (\emph{domain specific}) increases re-ID accuracy on real-world degraded images.}
    \label{fig:visual_abstract}
\end{figure}
Ultimately successful re-ID is strongly dependent on images of sufficiently good quality. In real-world scenarios, images often suffer from various degradation factors such as blur (e.g. motion or out-of-focus blur due to animal movement), downscaling (low resolution due to a large distance between the animal and the camera), noise, compression artifacts etc. All these significantly reduce the fine-scale discriminative information, negatively affecting the re-ID and eventually limiting the downstream ecological studies. In fact, it is \emph{precisely} these low-quality images that challenge both automated and human-based identity matching. Such images are often discarded at early stages of data curation, both when datasets are prepared for downstream ecological analyses and when assembling training data for deep feature models \cite{otarashvili_multispecies_2024}.
\emph{Understanding the effect of image degradation in wildlife re-ID and developing methods to boost and robustify associated models in the low-image quality regime is thus of great importance}.\\[0.5em]
\noindent
\textbf{Related work}\\[0.5em]
Previous related work is either focused on \emph{coarse-grained classification} or on \emph{human re-ID}, which, as we remark later, even though related, both differ from wildlife re-ID with respect to practical set-up, dataset characteristics, methods and general scope. 

With regards to coarse-grained classification, in \cite{pei_effects_2021, gade_impact_2023}, artificially degraded images were used to expand the training dataset of CNN classifiers, which were also evaluated on artificially degraded images. It was found that accuracy is highest when test images have the same type and degree of degradations as in training, and it decreases significantly when the discrepancy between training and testing degradations increases. CNNs can interpolate between known blur severities but struggle to extrapolate to stronger blur levels than those seen during training \cite{gade_impact_2023}.
Furthermore, models trained on degraded images performed worse on clean test images than non-augmented models. 
Finally, it was observed that applying image restoration algorithms did not improve the performance of classifiers significantly \cite{pei_effects_2021}. 
We mention already that the above trends might not directly translate to wildlife re-ID due to several intricacies of the latter:
\begin{enumerate}
\item [(i)] Wildlife re-ID is based on fine-grained information, relying on subtle discriminative details and localised patterns which might be particularly susceptible to degradations.
 Different species have different type of identifying patterns potentially making the susceptibility to degradations species-specific.
\item[(iii)] In contrast to coarse-grain classification where all classes are seen during training, in wildlife re-ID there are always new individuals in the database that are not represented in the training set and thus augmented training cannot be applied to those.
\item[(iv)] Real-world degradations in wildlife images are very diverse, species- and dataset-specific and largely unknown, leading to a domain gap between synthetic and real-world data.
\item[(v)] In contrast to coarse-grain classification, there do not exist wildlife datasets with annotation on the image quality that can facilitate testing re-ID in the low image quality regime.
\item[(vi)] Evaluation of wildlife re-ID differs to coarse-grained classification, based on a combination of Rank-$k$ accuracy, human verification effort and effects in the downstream ecological task \cite{stewart2021animal}.
\end{enumerate}
With regards to human re-ID, there is a larger number of works dealing with image degradation mostly focusing on combining re-ID with image restoration to mitigate degradation effects \cite{huang_real-world_2020, huang_learning_2022, cheng_low-resolution_2019, zheng_joint_2022}. Most existing works focus on a specific type of degradation, e.g.\ variation in illumination conditions \cite{bak_domain_2018}, different modalities \cite{wu_rgb-infrared_2017, chen_neural_2021}, or low-resolution \cite{wang_cascaded_2018, jiao_deep_2018, cheng_low-resolution_2019, zheng_joint_2022, wang_studying_2016}. 
Again, animal and human re-ID have many fundamental differences: 
\begin{enumerate}
\item [(i)] Human re-ID focuses exclusively on human subjects, whereas wildlife re-ID encompasses a wide range of species, each presenting distinct challenges in terms of appearance, morphology, and behaviour. 
\item[(ii)] Wildlife data are collected in different environments and are more difficult to gather, whereas most human re-ID data are captured in urban environments. Connected to that, animal poses are highly species- and environment-specific, while human datasets predominantly feature bipedal walking postures. 
\item[(iii)] 
Human re-ID typically focuses on tracking or matching individuals across non-overlapping cameras within the same encounter or a short time window, or across different times within the same camera view. In contrast, animal re-ID aims to recognise individuals across longer time spans and distinct encounters, which results in distribution shifts between images.
\item[(v)] 
 Human re-ID systems are often associated with privacy and surveillance concerns.
In contrast, animal re-ID is primarily developed for ecological research and wildlife conservation, serving as one of the most non-invasive techniques for studying and monitoring animals in their natural habitats.
\end{enumerate}
Overall, these differences with respect to human re-ID and general coarse-grained classification highlight the necessity to conduct research on understanding and addressing image degradations on animal re-ID as well, which as far as we are concerned has not been done so far in the literature.\\[0.5em]
\noindent
\textbf{Our main contributions are the following:}
\begin{enumerate}
\item [(1)] We first train a baseline deep feature extractor model across different species (18 datasets), and track its retrieval performance on artificially degraded query sets. We show that the adverse effects of degradations in animal re-ID are \emph{highly species/datasets specific}: 
Species/datasets for which the baseline model exhibits similar performance for non-degraded query images may show markedly different performance once these images are degraded.
\item[(2)] We introduce a \emph{model-based degradation-based augmented training framework} using three families of artificial degradations of increasing complexity (simple, diverse, diverse$^{+}$)
to generate synthetic degraded training images that simulate real-world conditions. 
We show that
this augmented training leads to an improved re-ID performance for query images corrupted with the same degradation type. 
However, in contrast to the coarse-grained classification, in datasets where the baseline model performs well (\textgreater80\% Rank-1 accuracy), the introduction of augmented training does not affect the re-ID performance on non-degraded query images. 
\item[(3)] Relevant to the real-world ecological setting where new individuals are added to the database all the time,
we show that our augmented training pipeline improves the re-ID performance even for degraded query images of individuals \emph{not seen} during training.
\item[(4)] We equip one of the datasets, SeaTurtleID2022 \cite{adam_seaturtleid2022_2024}, with image quality annotations provided by human re-ID experts specifically based on factors that affect the re-ID task. The lowest-quality subset constitutes the first animal re-ID benchmark based on \emph{real-world degraded images}, enabling systematic evaluation of re-ID models under low-image-quality conditions.
\item[(5)] We test the augmented models on this dataset and show that the introduction of diverse degradations during training leads to improved performance compared to the baseline model (gain of 8.5\% Rank-1 accuracy).
\item[(6)] Finally, to showcase the advantage of capturing  domain-specific degradations more accurately, we employ a style-transfer GAN to generate synthetic but realistic low quality samples that imitate the real-world ones from the annotated dataset  and use those for data augmentation. This results to even more superior re-ID performance compared to the previous handcrafted augmentation framework on the same dataset.
\end{enumerate}

\section{Degradation-based training data augmentation for wildlife re-ID}\label{degradation_ssection}

The idea behind degradation-based training data augmentation,
is to include in the training set images that have been subjected to the same type of degradations as the ones for which the model is targeted for \cite{mikolajczyk_data_2018, bak_domain_2018, pei_effects_2021}. Since real-world degradations are largely unknown, the degradations applied during training aim only to approximate these to a satisfactory degree.
This approximation has been shown to play a crucial role in coarse-grained classification \cite{mikolajczyk_data_2018, pei_effects_2021, gade_impact_2023} and, even more prominently, in image restoration. Models trained with image pairs (clean images plus degraded versions) have demonstrated substantial performance improvements in blind image restoration settings \cite{Wang_2021_ICCV, zhang_designing_2021}.

We adopt similar complex degradation pipelines for our training data augmentation. For reference, we start with the \emph{simple degradation} approach. Here, a low resolution 
image $\mathrm{z}$ is generated by first convolving the original image $\mathrm{x}$ with a Gaussian kernel $\mathrm{k}$, followed by a downscaling operation $D_s$ with scale factor $s$ and additive Gaussian noise $\epsilon$, mathematically modeled by
\begin{equation}\label{simple_degradation}
   \mathrm{z}=D_{s}(\mathrm{x} \ast \mathrm{k)}  \ + \epsilon.
\end{equation}
\noindent
Real-world degraded samples exhibit far more severe and complex distortions, and simple degradation methods like \eqref{simple_degradation} fail to reproduce those. Thus, in order to sufficiently augment training data for robust re-ID, we employ complex degradation pipelines involving several operations, discussed in Section \ref{sec:model_based_degradations}.

Alternatively, when focusing on specific domains and annotated good- and bad-quality images are available, the degradation process can be learned in a data-driven manner. To showcase this, we leverage on novel expert-provided quality annotations of the  SeaTurtleID2022 dataset, see Section \ref{Sec:real_world_deg_images}, and we train a style-transfer GAN to model the mapping between  two quality domains (from highest to lowest). This is then used to generate synthetic domain-specific low quality training samples for data augmentation, see Section \ref{sec:data_driven_degradations}.

\subsection{Model-based degradation details}\label{sec:model_based_degradations}

We describe the building blocks that  form our complex degradation pipelines (model-based). Details about the exact formulas and parameters are in the supplementary material.\\[0.0em]
\emph{\textbf{Blur:}}
The blurring operations used in our pipeline include \textit{Gaussian}, \textit{Generalised Gaussian}, \textit{Motion}, and \textit{Defocus Blur}. Gaussian blur is realised via convolution with a standard Gaussian kernel. Generalised Gaussian blur employs kernels that can range from box-like to very peaked blurs, simulating a wide range of isotropic and anisotropic blurs \cite{liu_estimating_2021}.
Motion blur is modelled as a convolution between the image and a motion kernel that represents the path of movement during exposure \cite{levin_understanding_nodate, brooks_learning_2019}. Finally, defocus blur is synthesised by combining a disc-shaped kernel and a Gaussian blur, which together simulate the shape of a camera’s aperture.\\[0.0em]
\emph{\textbf{Downscaling:}}
To simulate the downscaling process in our pipeline and to increase the diversity of degradations, we use three methods, namely bicubic, bilinear, and nearest-neighbour with a scaling factor of 2 or 4.\\[0.0em] 
\emph{\textbf{Noise \& artifacts:}}
Here we only use Gaussian noise, since in the absence of information on the noise distribution, it is the most appropriate choice \cite{park_gaussian_2013}. Finally, we apply JPEG compression to simulate compression artifacts.

\begin{figure*}[t]
    \centering
    \includegraphics[
    width=\textwidth]{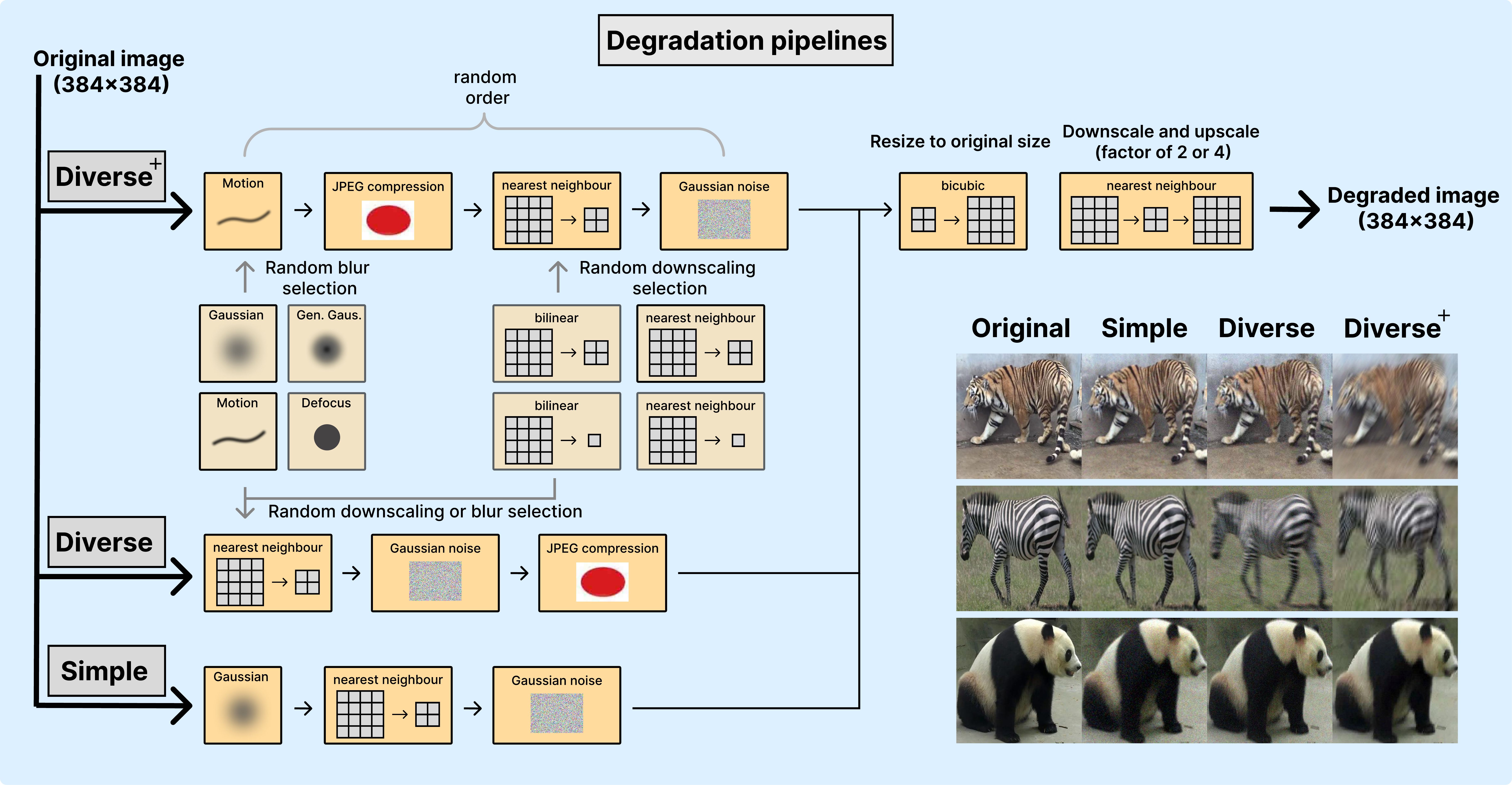}    
    \caption{The employed model-based degradation pipelines (simple, diverse and diverse$^{+}$) used for training augmentation, along with corresponding image samples.\vspace{-0.5em}
    }
    \label{fig:degradation_pipelines}
\end{figure*}

\vspace{0.5em}
\noindent
\textbf{Degradation pipelines: Diverse$^+$, diverse and simple:}
Our degradation pipelines are inspired by the ones employed in image restoration for creating paired clean and degraded images for training \cite{zhang_designing_2021}.
Each image undergoes a random sequence of the above degradations with randomly chosen parameters, see supplementary material for details. We adopt two such pipelines, which we name \emph{diverse} and \emph{diverse$^+$}. The output is always a degraded image with resolution compatible with the input dimensions of our re-ID models (384$\times$384). We note that this is also the resolution of the input images to the pipeline, so we have the same starting point for all the models.

In the \textbf{diverse$^+$ degradation pipeline}, the input image is subjected to four main degradation operations (blur, downscaling, Gaussian noise, JPEG compression) applied in a random order. The blur operation consists of an application of one out of four blurs described previously, with randomly chosen corresponding parameters. The downscaling operation consists of one of four types of downscaling depending on the factor (2 or 4) and the method (bilinear or nearest neighbour). After these four degradation operations, the image is resized back to its original size via bicubic interpolation. Finally, a successive downscaling and upscaling using the same factor (2 or 4) and nearest neighbour is performed in order to simulate post-degradation resampling.
In the \textbf{diverse degradation pipeline}, only one out of the eight possible blur or downscaling degradations described above is applied to the input image, followed by Gaussian noise and JPEG compression. For reference, we also define the \textbf{simple degradation pipeline} as in \eqref{simple_degradation}. There, only a Gaussian blur is applied followed by downscaling (with factor 2 or 4) and Gaussian noise. Both of the latter pipelines end by 
 resizing to the original size and the downscaling and upscaling procedure as in the last step of the diverse$^+$. All degradation pipelines are visualised in Figure \ref{fig:degradation_pipelines} together with some sample images.

\subsection{A re-ID dataset of real-world degraded animal images}\label{Sec:real_world_deg_images}

\begin{figure}[t]
    \centering
    \includegraphics[width=1\textwidth]{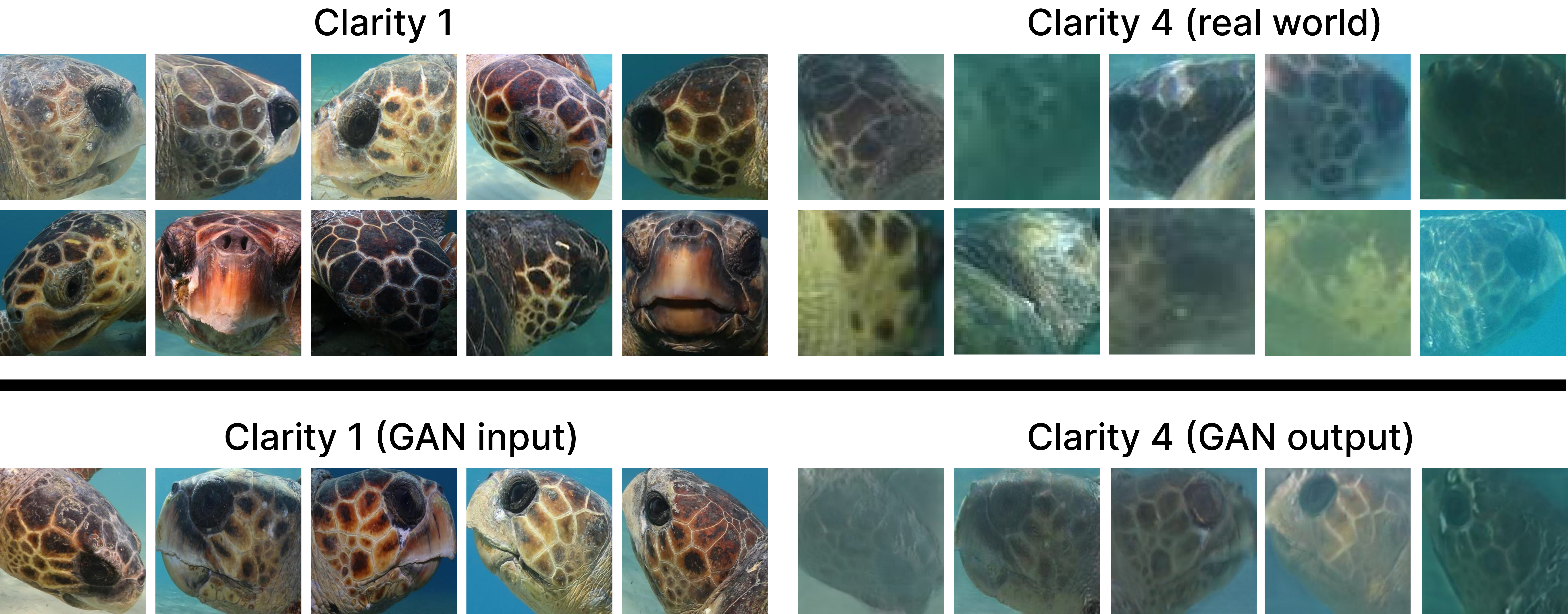}      
    \caption{Top: Samples from the expert-annotated images of the SeaTurtleID2022 dataset \cite{adam_seaturtleid2022_2024}   with respect to clarity factors affecting re-ID. Left: ``Clarity 1''-images, where identifying details like facial scales (if present in the image) are clearly visible. Right: ``Clarity 4''-images, where such details are hardly visible, due to factors like blur, low resolution, distortion artifacts, etc. Bottom: Simulated ``Clarity 4'' images via a style-transfer GAN (input and output images), for domain-specific degradation-based augmented training.
    }
    \label{fig:clarities}
\end{figure}

In order to evaluate the augmented models on real-world conditions, it is necessary to have a dataset with image quality annotations. For this, we used the SeaTurtleID2022 dataset \cite{adam_seaturtleid2022_2024} (heads only) which consists of underwater images of loggerhead sea turtles, each of which can be identified by its unique facial scale pattern. Every image was given a clarity score (1, 2, 3 \& 4) by sea turtle re-ID experts, with 1 being the best clarity and 4 the worst. The annotation was based on clarity factors that specifically affect the re-ID task from the human perspective, e.g. based on whether the face details were deemed clear enough for manual re-ID. The annotation was done so that it does not depend on head orientation but only on the ability of the human to easily see image details. Out of 7582 images, 1514, 2346, 2535 and 1187 were given clarity scores equal to 1, 2, 3 and 4, respectively.
In order to minimise subjective judgement, and have clear differences in image quality, we focus here on the clarity 1 and 4 groups as representative opposing examples of ``very easy'' and ``very difficult'' images for re-ID, see Figure \ref{fig:clarities}(top) for some samples. In particular, the clarity 4 group constitutes the first in our knowledge animal re-ID dataset with real-world degraded images. It consists of images with low resolution (animals far away from the camera), underwater distortion, illumination artifacts, blur due to movement, etc.

\subsection{Learning-based degraded data augmentation}\label{sec:data_driven_degradations}
Given the  good- and bad-quality images described in Section \ref{Sec:real_world_deg_images}, i.e.  clarity 1 and 4, we train a style-transfer GAN \cite{park_contrastive_2020} to learn the mapping between the clarity 1 (best quality) and clarity 4 (worst quality) domains. To preserve identity, the training objective is augmented with a re-ID loss. We provide details  in the supplementary material. The trained model is then used to generate synthetic clarity 4 images, which are incorporated into the re-ID training set for domain-specific degradation-based augmented training. In Figure \ref{fig:clarities}(bottom) we  provide some output samples of the GAN (simulated clarity 4 images) and the corresponding input images.

\section{Formulation of experiments}\label{Sec:Datasets}

\subsection{Datasets}\label{Sec:Datasets}
\begin{figure}[t]
\begin{minipage}[c]{0.25\textwidth}
\resizebox{0.99\textwidth}{!}{
\begin{tabular}{@{}l@{\hspace{5px}}r@{\hspace{8px}}r}
\toprule
 & \textbf{images} & \textbf{ids} \\
\midrule
\midrule
ATRW (tigers) \cite{li_atrw_2020}   & 5,415 & 182   \\ 
BelugaID \cite{belugaid}   & 8,559 & 788  \\
CTai (chimpanzees) \cite{freytag2016chimpanzee}   & 4,662 & 71  \\
ELPephants \cite{korschens_elpephants_2019}				 & 2078 & 274 \\
Giraffes \cite{miele_revisiting_2021}   & 1,368 & 178   \\
GiraffeZebraID \cite{parham2017animal}   & 6,898 & 2,051 \\
HumpbackWhaleID \cite{cheeseman_advanced_2022}  & 15,697 & 5004 \\
HyenaID2022 \cite{botswana2022}   & 3,129 & 256 \\
IPanda50 \cite{wang_giant_2021}   & 6,874 & 50  \\
LeopardID2022 \cite{botswana2022}   & 6,806 & 430  \\
MacaqueFaces \cite{witham_automated_2018}  & 6280 & 34 \\
NyalaData \cite{dlamini_automated_2020}   & 1,942 & 237  \\
OpenCows2020 \cite{andrew_visual_2021}   & 4,736 & 46  \\
SealID \cite{nepovinnykh_sealid_2022}   & 2,080 & 57  \\
SeaTurtleID2022 \cite{adam_seaturtleid2022_2024}   & 8,729 & 438   \\
StripeSpotter \cite{lahiri_biometric_2011}   & 820 & 45  \\
WhaleSharkID \cite{holmberg_estimating_2009}   & 7,693 & 543  \\
ZindiTurtleRecall \cite{zinditurtles}   & 12,803 & 2,265  \\
\midrule
TOTAL & 106,569 & 12,949 \\
\bottomrule
\end{tabular}
}
\end{minipage}\hspace{0.25cm}
\begin{minipage}[c]{0.73\textwidth}
\includegraphics[width=\textwidth]{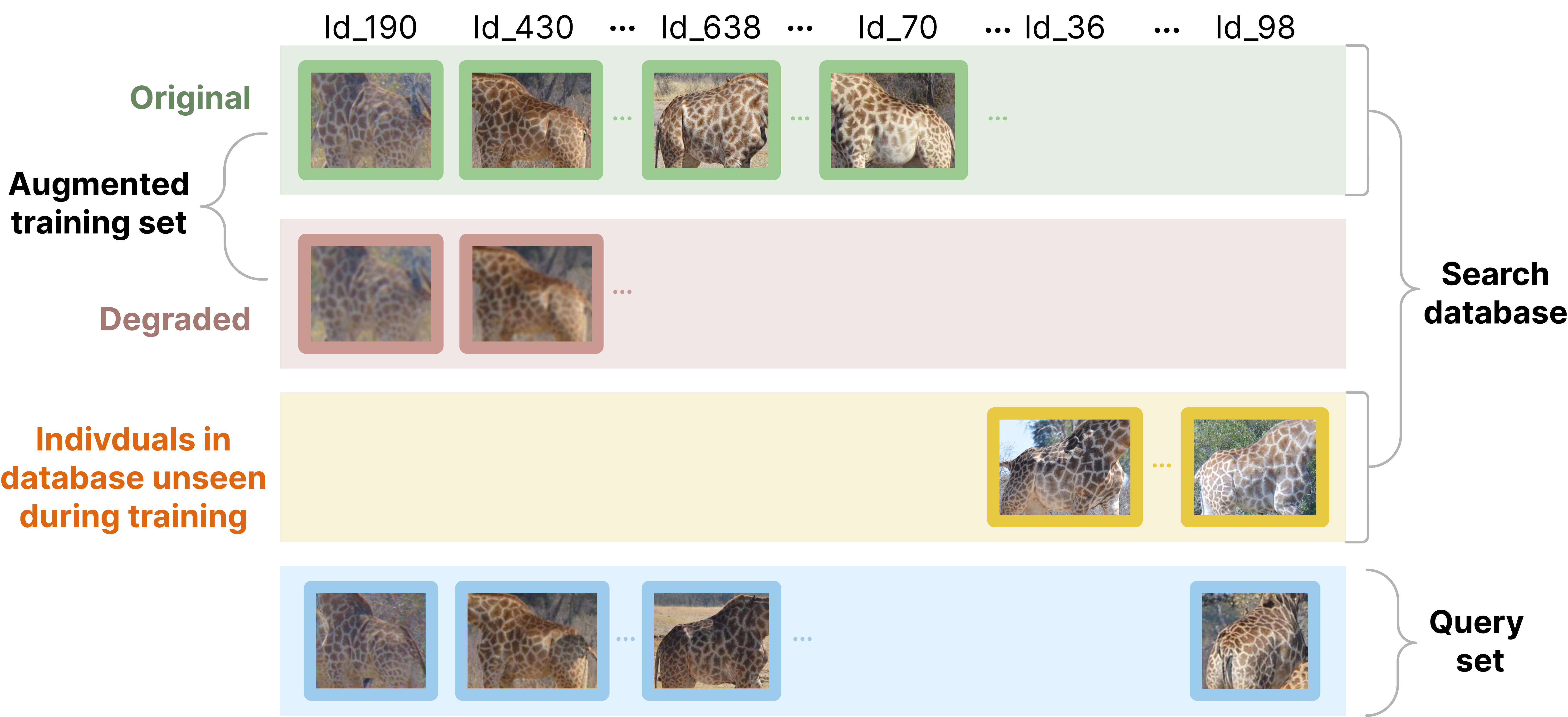}
\end{minipage}
\caption{Datasets used for training and testing our models and the corresponding split configuration into training, search database and query sets. A part of the training set is augmented using degradations, and a subgroup of individuals is not included in the training set to test the generalisability of the models.
}
\label{fig:datasets_split}
\end{figure}

The datasets that we used are publicly available from the WildlifeDatasets library \cite{cermak_wildlifedatasets_2024, adam_wildlifereid-10k_2024}. 
 We selected a subset of this collection, excluding datasets that were deemed unsuitable for re-ID experiments (e.g.\ flies) and avoiding too many datasets of the same species. Overall, 18 datasets were used, with a total of approximately $13k$ individuals and $100k$ images, see Figure \ref{fig:datasets_split}(left) for a full list. We note that these datasets are used to test only the models trained via model-based degradation augmentations, whereas the GAN-based (domain-specific) is tested on its domain, that is, the  SeaTurtleID2022 dataset.
 
\subsection{Data splitting procedure}\label{Sec:Prop_Data_Split}

In our setup, we employ a structured data partitioning strategy that enables a  comprehensive evaluation of the degradation-based training data augmentation, which is visualised in Figure \ref{fig:datasets_split}(right). Here we focus on a \emph{closed-set} setting between the query set and the database, meaning that in all the experiments the query images correspond to animals already present in the database. 
We note that this does not capture the cases of newly introduced individuals (\emph{open-set} setting) which is most the times the case in real-world scenarios. However we emphasise that our aim here is mainly on increasing the probability of retrieval of low quality images rather than focusing on the open-set setting which is a challenging problem on its own right, however of different nature.

Every dataset is divided into a subset $\mathrm{D_{seen-ids}}$ consisting of individuals that have images included in the training set, and a subset $\mathrm{D_{unseen-ids}}$ consisting of individuals not seen at all during training. This allows us to evaluate the generalisability of the models through their evaluation on identities unseen during training. The subset $\mathrm{D_{seen-ids}}$ is split into (i) a group of images that are included in the training set and also in the search database of the retrieval experiments and (ii) a group of images that are part of the query set of the retrieval experiments. The subset $\mathrm{D_{unseen-ids}}$ is split into (i) images that are part of the search database and (ii) images that are part of the query set.

The training subgroup is dynamically augmented with images gone through one of the three degradation pipelines in an online manner. We note that the degraded images in the training set are not part of the search database. Exact numbers of images and individuals of every dataset assigned to the above groups can be found in the supplementary material.

\subsection{Model architecture}
For augmented models using model-based degradations, we use   a Swin Transformer architecture \cite{liu_swin_2021} as a  backbone, which is the same as the MegaDescriptor architecture \cite{cermak_wildlifedatasets_2024}. However, unlike MegaDescriptor, we use CurricularFace loss \cite{huang_curricularface_2020}, an adaptive softmax- and margin-based loss that dynamically adjusts the decision boundaries according to the training stage and sample difficulty. Implementation details, including the training procedure and hyperparameter settings, are provided in the supplementary material.

Experiments involving learning-based data augmentation employ an EfficientNetV2 backbone \cite{pmlr-v139-tan21a}, which has also been shown to be effective for wildlife re-ID \cite{otarashvili_multispecies_2024}. This choice substantially reduces the computational cost of training while demonstrating that the proposed augmentation is compatible with a different re-ID architecture. Here, we follow the training protocol of \cite{otarashvili_multispecies_2024} which involves a sub-center ArcFace loss.

\subsection{Evaluation Metrics}
To evaluate the performance of the re-ID models, we use the  
Rank-$k$  accuracy which denotes the probability that a correct match appears within the top-$k$ ranked retrieved results. We report Rank-$k$ values at $ k\in\{1,5,10,20\}$. 
We also use the mean Average Precision (mAP) metric \cite{zheng_scalable_2015} which measures the overall retrieval performance when there are multiple positive matches.

\subsection{Augmented re-ID models}\label{Sec:Exp_formulation}
We first train a baseline transformer-based model on the 18 datasets mentioned in Section \ref{Sec:Datasets}, by following the splitting procedure detailed in Section \ref{Sec:Prop_Data_Split}. For this baseline we do not employ degradation-based training data augmentation, but we only apply standard geometric and color augmentations via RandAugment \cite{Cubuk_2020_CVPR_Workshops}. 
Subsequently, we train three models on the same datasets and data split, by employing the simple, diverse and diverse$^{+}$ degradations for augmented training. 
We hypothesise that the degradations in the simple model are not rich enough to capture the complexity of the real-world ones, and thus we expect the diverse and diverse$^{+}$ models to perform better in the real-world low quality regime, e.g.\ the clarity 4 subset of the SeaTurtleID2022 dataset.

In the first set of experiments, the query set is itself artificially degraded using the diverse and diverse$^{+}$ pipelines and we track down the performance of each model, separately for seen and unseen individuals during training as well as for every species/dataset. Even though this is not a ``real-world'' experiment, it informs us about each model’s robustness and generalisation capability under degradation and species-specific differences.

In the second set of experiments, the models are evaluated on the expert-annotated SeaTurtleID2022 dataset \cite{adam_seaturtleid2022_2024}, tracking down their performance on the subsets of different clarity scores, with the clarity 4 subset serving as a benchmark for real-world degraded images. There, 
we additionally evaluate the proposed learning-based augmentation on the SeaTurtleID2022 dataset. Using the same training split and evaluation protocol, we train an  EfficientNetV2 model using the diverse degradation pipeline further augmented with GAN-generated images.


\section{Results}

\begin{figure}[t]
\begin{minipage}[c]{0.35\textwidth}
    \centering
    \includegraphics[width=1\textwidth]{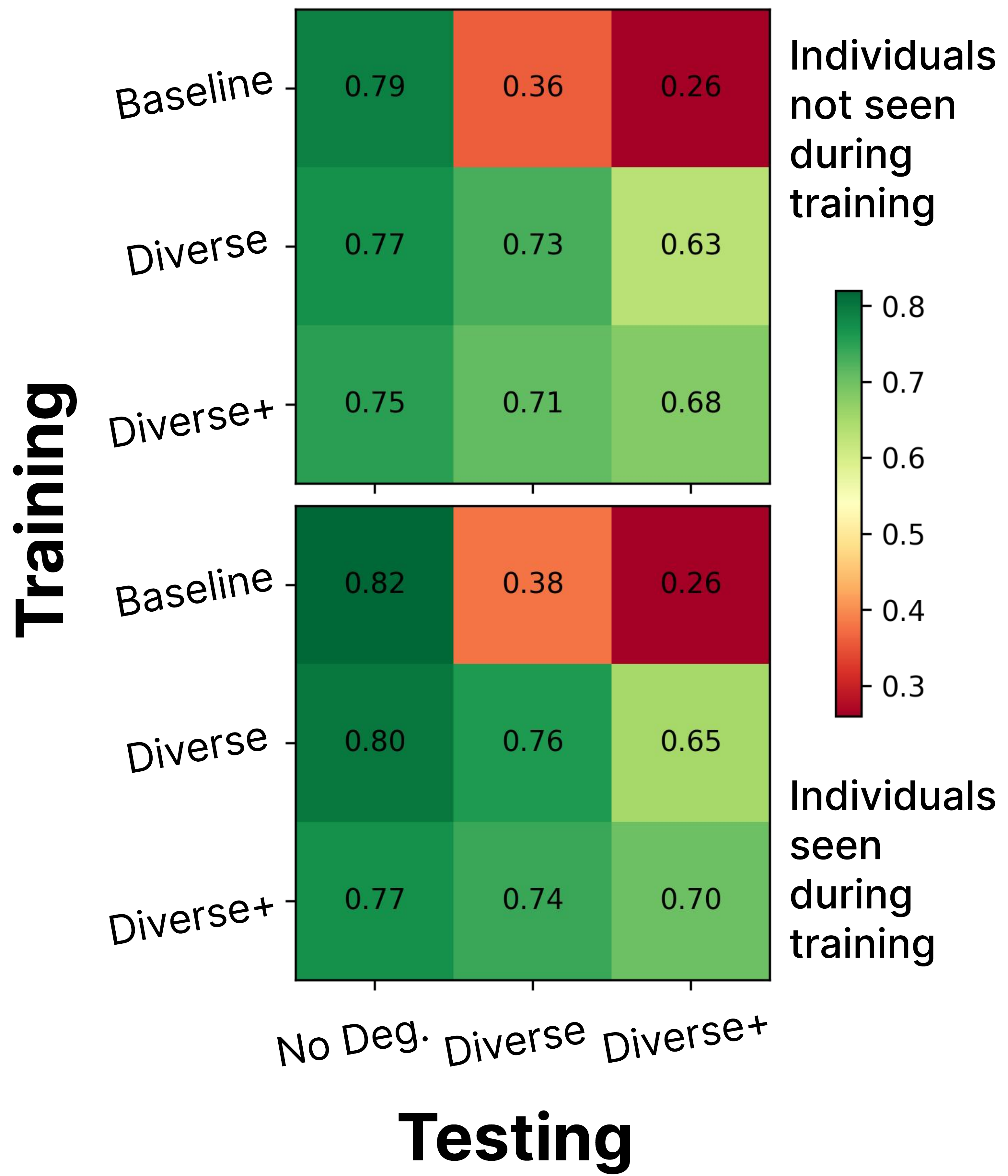}
\end{minipage}
\begin{minipage}[c]{0.64\textwidth}
    \centering
    \vspace{-1.5em}
    \includegraphics[width=0.99\textwidth,trim={0 0 1.0cm 0},clip]{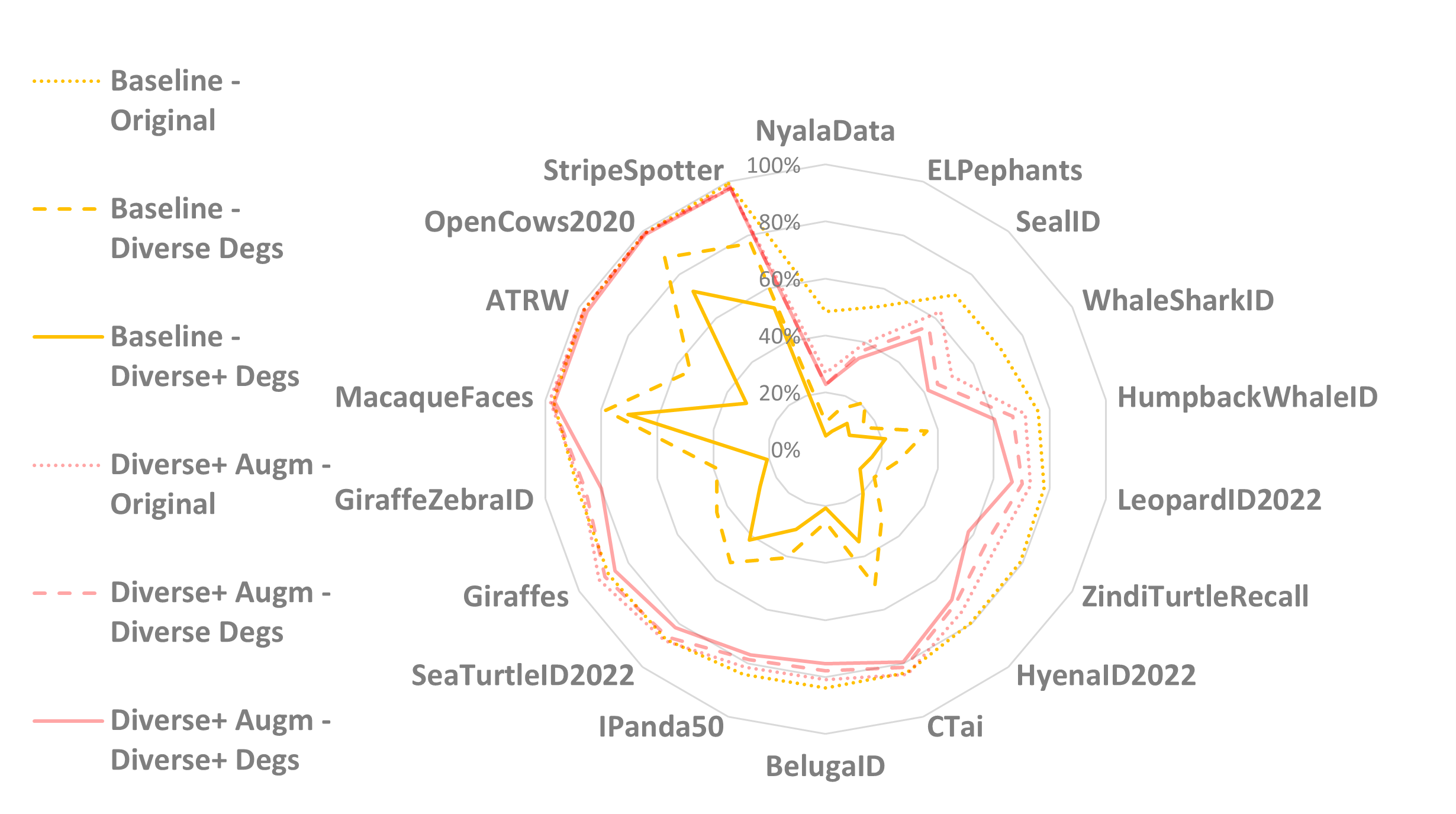}
\end{minipage}   

    \caption{Left: Rank-1 accuracies of the baseline model, and the models with diverse, and diverse$^{+}$ augmentations on their training set, evaluated on the original non-degraded query set and the two artificially degraded ones (diverse and diverse$^{+}$). Accuracies are averaged for all 18 datasets and reported separately for individuals seen and not seen during training. Right: Rank-1 accuracies for the same experiments reported separately for each of the 18 datasets, where we observe that the magnitude of the adverse effects of image degradations on re-ID performance is species-specific.
    }
    \label{fig:results_arti_degs}
\end{figure}

\subsection{Effects of augmented training on artificially degraded query sets}

In Figure \ref{fig:results_arti_degs} (left) we report the performance (Rank-1 accuracy; averaged for all 18 datasets) of the baseline, diverse and diverse$^{+}$ models, for query sets that are non-degraded, and degraded with diverse and diverse$^{+}$ degradations. This results in overall 9 re-ID experiments. Rank-1 accuracies are reported separately for those individuals seen and not seen during training.\\[0.2em]
\textbf{General robustness under degradations}: Upon fixing a model, 
performance is decreasing when moving from the non-degraded to diverse and to diverse$^+$ degraded query set, indicating that the latter are the most detrimental degradations to the re-ID task. Both augmented models perform much better in degraded query sets than the non-augmented baseline. For each of the diverse and diverse$^{+}$ degraded query sets, the model that performs the best is the one trained with the same type of degradations. However, the diverse$^{+}$ model performs almost equally well with the diverse one, also in the diverse-degraded query set.\\[0.2em]
\textbf{Robustness for individuals not seen during training:} 
Importantly, augmented models provide robustness against (here artificial) degradations also for individuals not seen during training. 
Note that Rank-1 accuracies for these individuals are only 2-3\% lower than the ones than corresponding to individuals seen during training.\\[0.2em] 
\textbf{Species-specific differences}: In Figure \ref{fig:results_arti_degs}(right), we show the Rank-1 accuracies for the same experiments, separately for each of the 18 datasets, where we have omitted the diverse model for visualisation sake, see supplementary material for the full graph. These are ordered clock-wise with respect to increasing accuracy for the baseline/non-degraded query set case (yellow dotted line).
While we observe similar trends to Figure \ref{fig:results_arti_degs}(right), we see that the effect of degradation to re-ID accuracy is highly species/dataset specific. Datasets for which the baseline model performs similarly for non-degraded query images, can show markedly different performance once these images are degraded; compare the yellow dotted lines with the dashed and solid yellow ones in Figure \ref{fig:results_arti_degs}(right). For instance, OpenCows2020, MacaqueFaces, ATRW, and GiraffeZebraID have all Rank-1 baseline accuracy of more than $90\%$ in the non-degraded query set, which drops to less than $32\%$ for the latter two but only to about $70\%$ for the former two, for diverse$^{+}$-degraded query sets. These species-specific differences tend to diminish for the augmented models, showing that their benefit is more uniform across datasets.\\[0.2em]
\textbf{Performance of the augmented models for non-degraded query sets}: Overall, we observe a slight performance drop for the augmented models on the non-degraded query set, see the first columns of the grids of Figure \ref{fig:results_arti_degs}(left). However, Figure \ref{fig:results_arti_degs} suggests that in datasets where the baseline model performs well (>$80\%$ Rank-1 accuracy), the introduction of augmented training does
not significantly affect the re-ID performance on non-degraded query images and in certain cases it even slightly increases it. This is in contrast to the coarse-grained classification where models trained on degraded images perform worse on non-degraded ones \cite{pei_effects_2021}.

\begin{figure*}[!ht]
 
\begin{minipage}[t]{0.47\textwidth}  
   \centering
   \textbf{Time-unaware, all individuals}
\end{minipage}
\begin{minipage}[t]{0.47\textwidth}  
   \centering
      \textbf{Time-aware,   not seen in training}
\end{minipage}
\vspace{0.2em}

\begin{minipage}[t]{0.02\textwidth} 
\vspace{-2.5cm}
\rotatebox{90}{Clarity 1}\\
\vspace{1.5cm}
\rotatebox{90}{Clarity 4}\\
\vspace{1.1cm}
\rotatebox{90}{Full query set}
\end{minipage}
\begin{minipage}[t]{0.45\textwidth}   
\centering
    \includegraphics[width=0.99\linewidth]{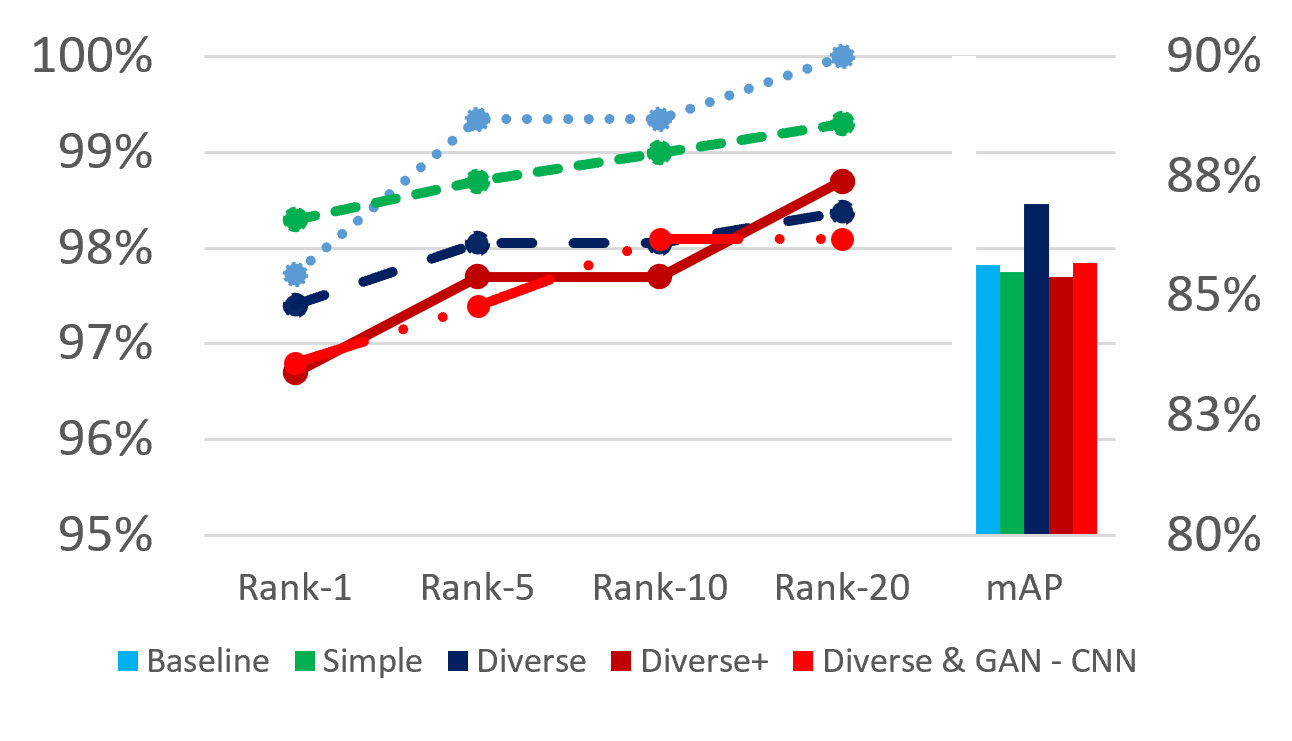}\\
    \includegraphics[width=0.99\linewidth]{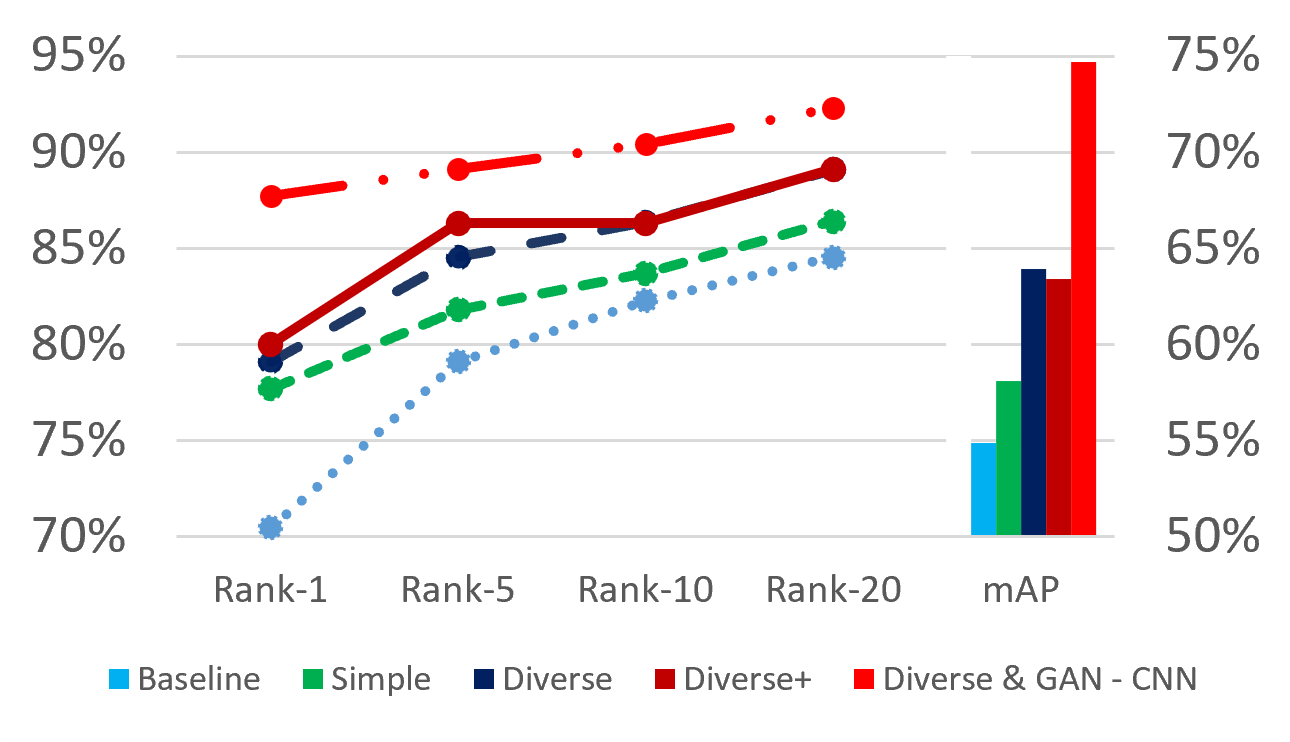}\\
    \includegraphics[width=0.99\linewidth]{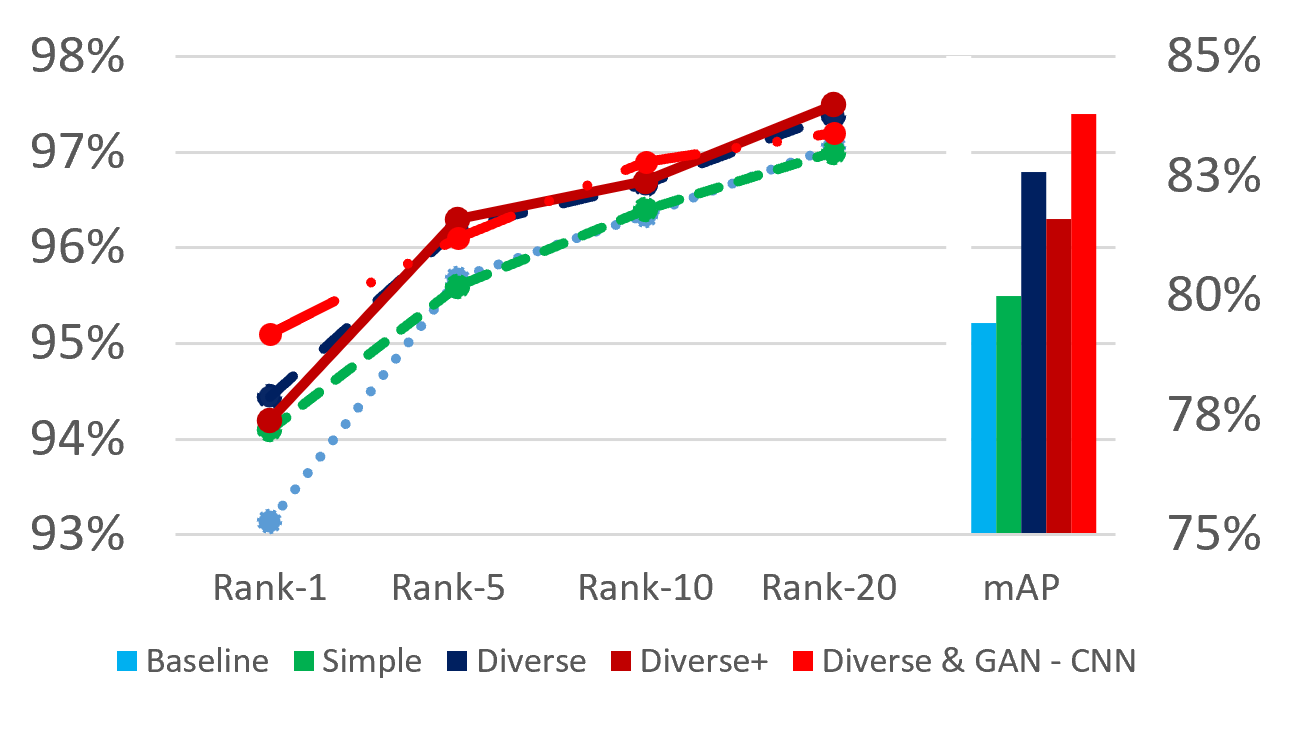}
\end{minipage}
\begin{minipage}[t]{0.45\textwidth}  
\centering
     \includegraphics[width=0.99\linewidth]{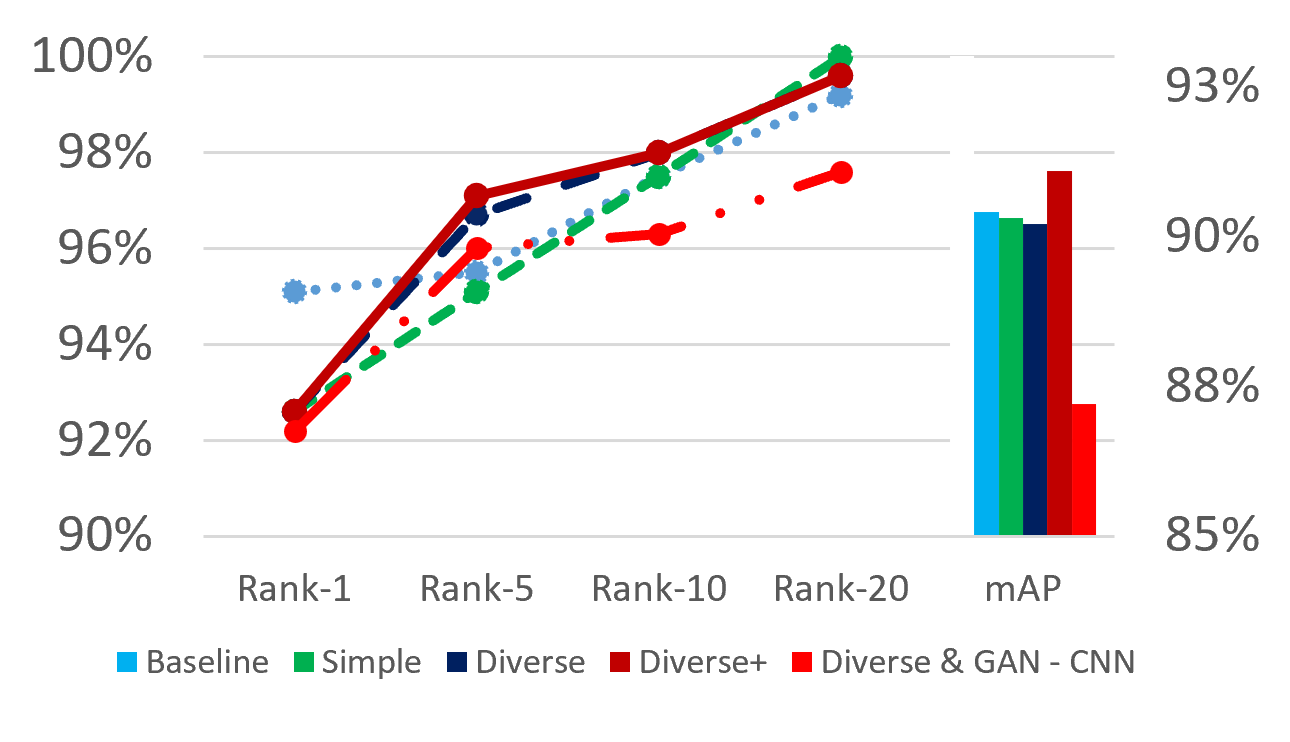}\\
    \includegraphics[width=0.99\linewidth]{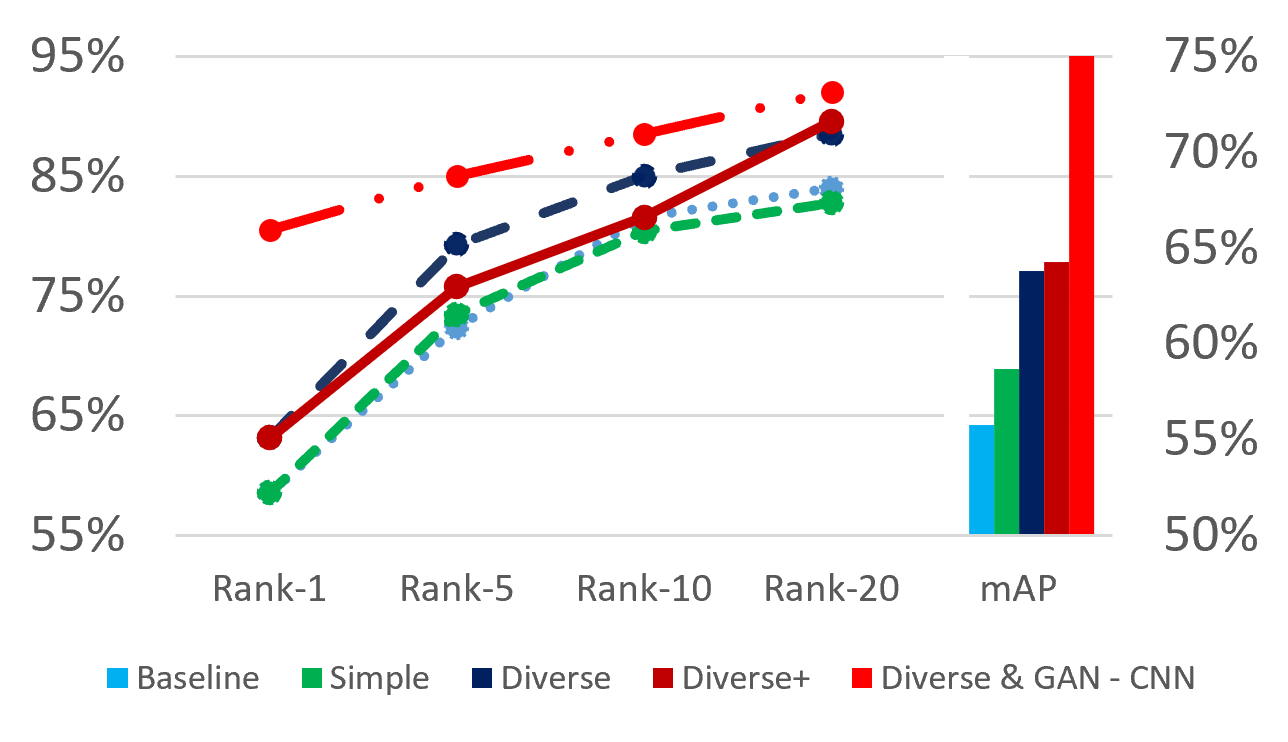}\\
    \includegraphics[width=0.99\linewidth]{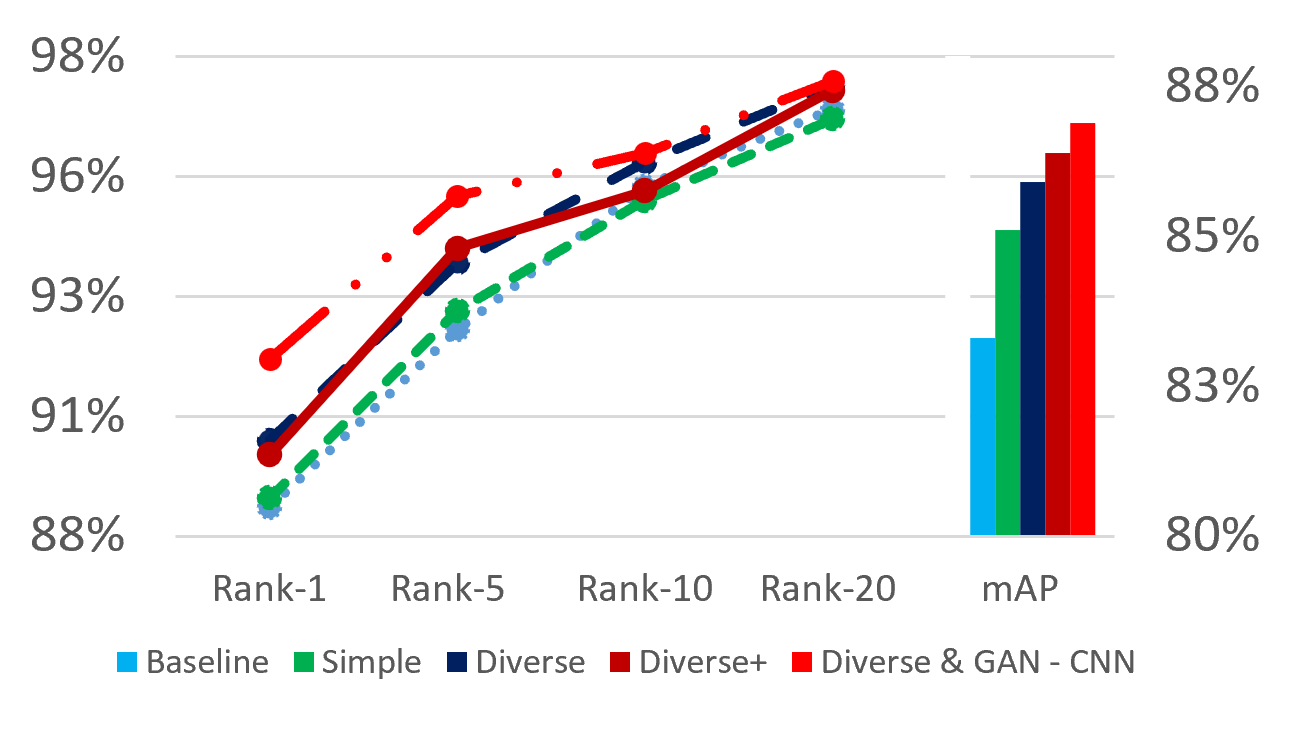}
\end{minipage}


\begin{minipage}[t]{0.49\textwidth}  
\centering
     \includegraphics[height=2.8cm]{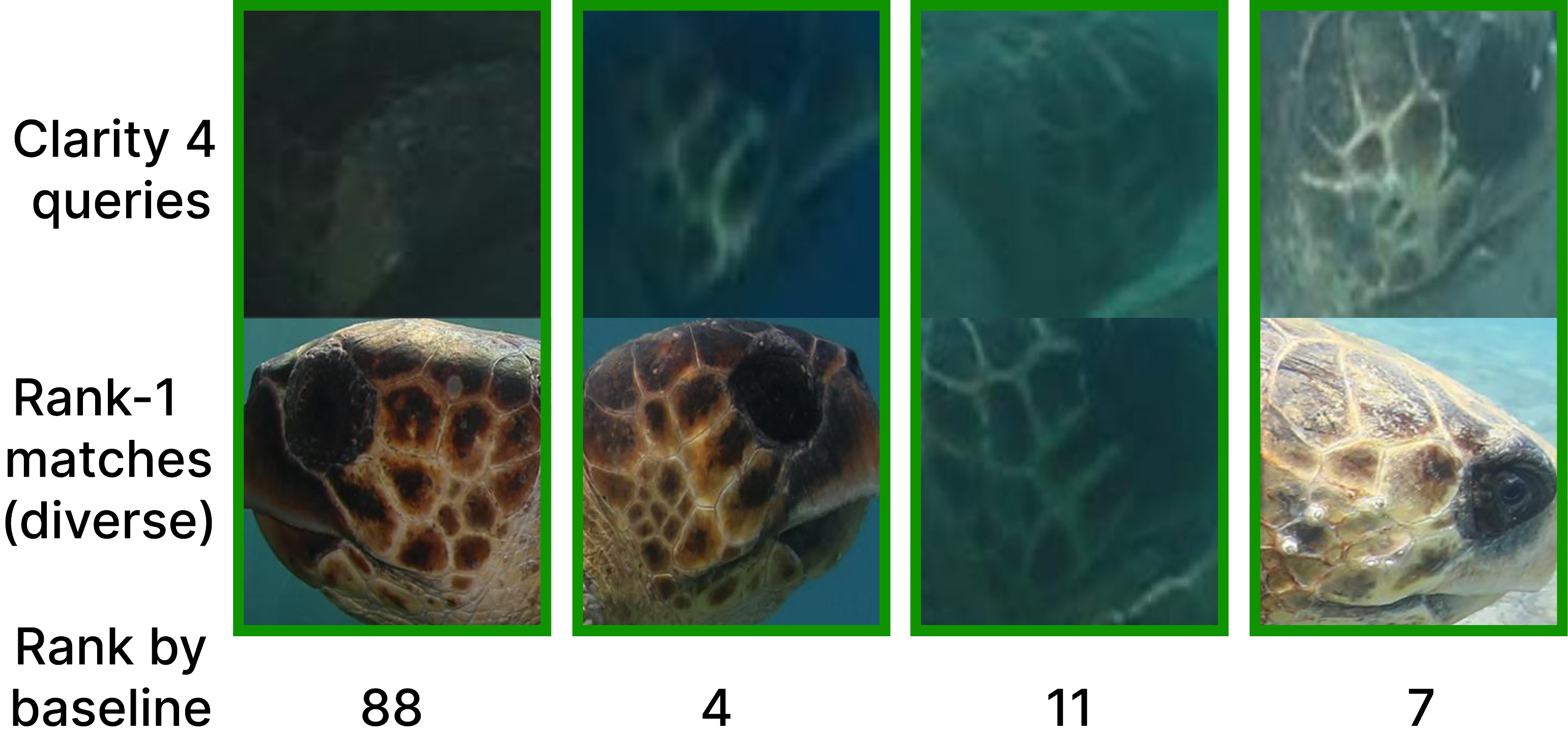}\\
\end{minipage}\hspace{-0.2cm}
\begin{minipage}[t]{0.49\textwidth}  
\centering
     \includegraphics[height=2.8cm]{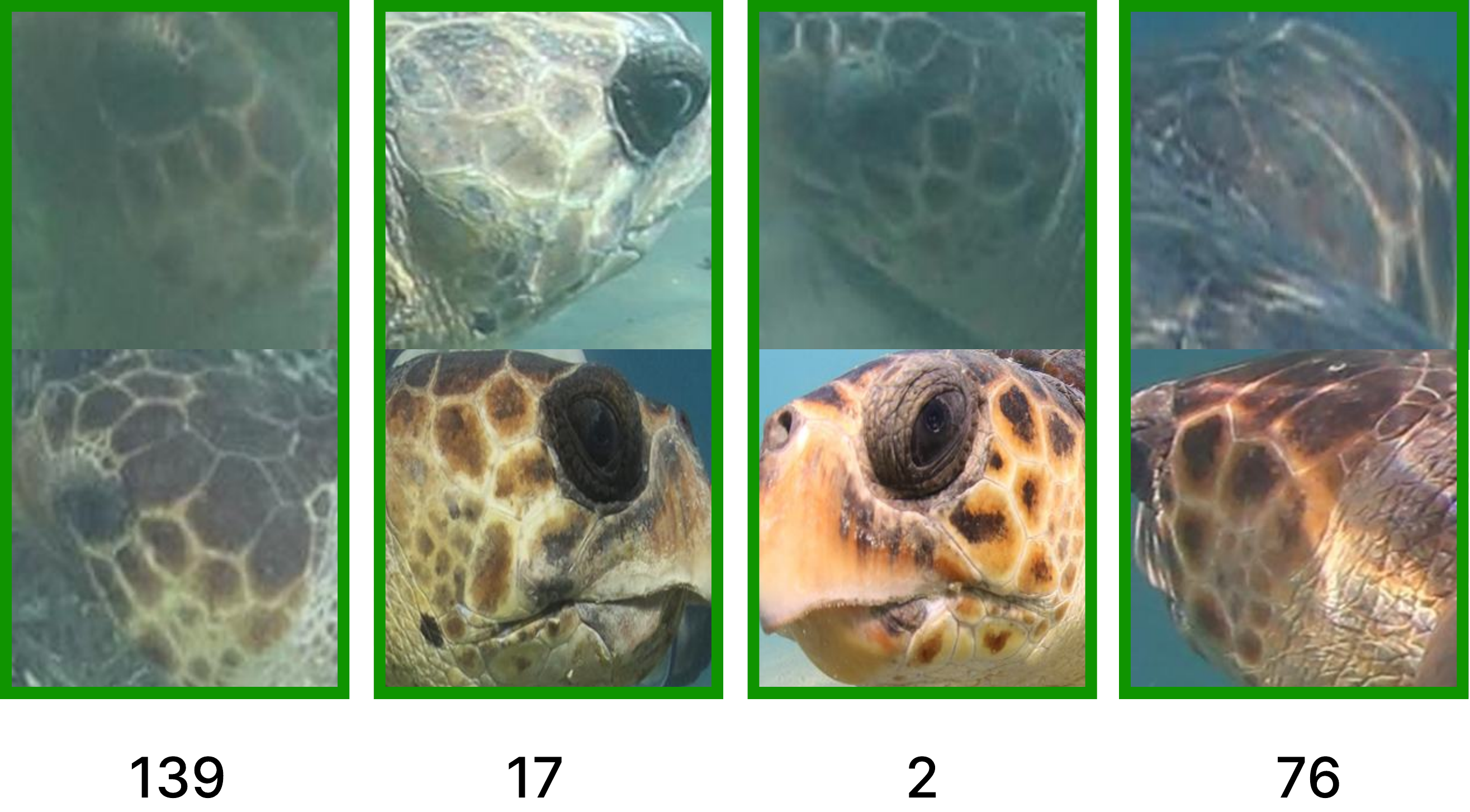}\\
\end{minipage}

    \caption{
    Top: Re-ID performance for query images of clarity 1, 4 and the full query set of SeaTurtleID2022. 
    In the right column, the query set is restricted to individuals not seen during training and whose images are split into database and query in a time-aware fashion. The augmented models have an overall better performance, especially the learning-based augmented model (``Diverse \& GAN-CNN''), which is particularly pronounced on the low quality, clarity 4 set. Bottom: Sample of correct Rank-1 retrieved matches of the diverse model (bottom images) to clarity 4 queries (top images) missed by the baseline model for both time-unaware and time-aware experiments. The numbers at the bottom denote the rank of the first correct retrieved image by the baseline model. Unlike the non-augmented baseline, the augmented model is able to match some images of extremely poor quality. \vspace{-1.5em}
    }
    \label{fig:TurtleHeads_clarity}
\end{figure*}

\subsection{Re-ID performance on real-world degraded images}

Having established robustness to artificial degradations, we next evaluate whether this capability persists when the augmented models are applied to real-world degraded query images. To this end, we assess the models on the clarity-annotated SeaTurtleID2022 dataset, and here we additionally report results for the learning-based augmented  re-ID model. 

In Figure \ref{fig:TurtleHeads_clarity}(left) we present the mAP and Rank-$k$ accuracies computed separately for query images of clarity 1 and 4, as well as for the full query set. 
In Figure \ref{fig:TurtleHeads_clarity}(right), we do the same but restricting the query set to individuals not seen during training, whose images have been re-split into a database and query set in a \emph{time-aware} fashion. This constitutes the typical scenario where the search database consists of images captured at an earlier time-point than the query ones, avoiding artificial inflation of re-ID performance \cite{adam_seaturtleid2022_2024}. \\[0.2em]
\textbf{Clarity 4 query images}: There is a clear improvement of re-ID performance of the augmented models on the clarity 4 query images over the baseline, about $8.5\%$ increase of Rank-1 accuracy for the diverse$^{+}$ model (time-unaware split) and an additional $\sim 7$\% for the  learning-based augmented model. Similar improvements are observed in the time-aware split, where the learning-based augmented model achieves a  $\sim 20\%$ increase of Rank-1 accuracy over the baseline. 
Similar results hold for clarity 2 and 3 images, see supplementary material.\\[0.2em] 
\textbf{Clarity 1 and full set of query images}: When restricting to clarity 1 query images, while the augmented models perform slightly worse than the baseline for the time-unaware split (but not always with respect to mAP), this is not the case for the time-aware split where performance is more balanced. For the full query set, the performance of the augmented models is higher than the baseline both with respect to mAP and Rank-$k$ accuracies, with the learning-based augmented model achieving overall the best performance. 

A detailed comparison of CNN-based re-ID models is provided in the supplementary material, where it is shown that for clarity 4 images and overall, the model augmented with learned degradations outperforms the one using only model-based ones. There we also show that the augmented models perform on par or even outperform  current state-of-the-art multi-species re-ID models such as MegaDescriptor \cite{cermak_wildlifedatasets_2024} on datasets that were not seen during training by any model. However, we stress that our focus is to show the benefit of degradation-based augmented training rather than introducing a new foundation model.
 \\[0.2em]
\textbf{Matching samples}: In Figure \ref{fig:TurtleHeads_clarity} we also provide examples of Rank-1 retrieved matches of the diverse model (bottom images) to clarity 4 queries (top images), that are missed by the baseline model. Each number at the bottom denotes the rank of the first retrieved image of this individual given by the baseline model. We observe that the augmented model is able to match images of extremely poor quality. In certain cases, see the last pair, opposite profiles are matched, which is a well-known property of deep-feature re-ID models for this taxon, see \cite{Adam_2025_side_similarities}.

\vspace{-0.5em}
\section{Conclusions}
\vspace{-0.1em}
Low-quality degraded images of wild animals with little discriminatory information challenge re-ID to the point that they are often discarded early in the re-ID workflow. Here, we showed that a degradation-based data augmentation framework can enhance the robustness of re-ID models in such image conditions. 

Similar to image restoration, we showed that it is crucial to augment the training set with good enough approximations of the real-world degradations. This was manifested in the increased performance of the diverse models over the simple one, which fails to capture the complexity of real-world degradations. However, despite both diverse and diverse$^{+}$ being beneficial, it was not clear if one would be preferable over the other. For example, the additional degradation complexity of diverse$^{+}$ over diverse did not always provide a clear performance edge. 
The learned degradation approach tailored to the target dataset yielded the largest performance improvements, suggesting that learning dataset-specific degradation models is a promising direction for developing more robust wildlife re-identification models, similarly to what has been observed for coarser-grained ecological tasks such as image classification \cite{beery2020synthetic, das2021domain}.

We anticipate that our introduction of the real-world clarity-annotated dataset will further boost animal re-ID research in the low-quality regime, but certainly, more such annotated datasets for various species are needed. We stress that automatic quality-annotation e.g.\ based on blur metrics might not be enough to produce them though, since as we showed, the effect of degradation to re-ID is species-specific. Thus, human expertise might be required to annotate exactly those types of images that constitute a bottleneck for re-ID for a given species.

In summary, our results suggest that there is a clear benefit in performing degradation-based augmented training, especially when dealing with low-quality images, and it should be considered during algorithmic development of animal re-ID methods.\\[0.2em]
{\small \textbf{Acknowledgements:} TP was supported by EPSRC. LA was supported by the Center for Advanced Machines and Manufacturing Technology (CAMAT)-TN02000028. This research used Queen Mary's Apocrita HPC facility, supported by QMUL Research-IT.}
\normalsize





\bibliographystyle{splncs04}
\bibliography{references-2}

\section{Supplementary Material}

\subsection{Degradation details}
\subsubsection{Model-based degradations}
We provide formulas and specifications for the degradations described in Section~$2.1$. All transformations were implemented using the Albumentations library \cite{2018arXiv180906839B}.
\\
\textbf{\emph{Blurring kernels:}}\\
- \textbf{Gaussian blur} is defined as the convolution of an image with a Gaussian kernel $\mathrm{k}^{(GB)}$ of size $k_s^{(GB)}\ =\ (2t+1)\times(2t+1)$. Formally, for each offset $(x,y)\in[-t,t]\times[-t,t]$, the Gaussian blur kernel is defined as 
\begin{equation}
   \mathrm{k}^{(GB)}(x,y) = \frac{1}{N}
   \exp\!\left(
   -\frac{1}{2}
   C^{\mathsf{T}} R^{\mathsf{T}}(\theta)\, \Sigma^{-1}\, R(\theta)C
   \right),\qquad C=[x,y]^{\mathsf{T}},
\end{equation}
where $(x,y)$ denote the spatial offsets (in pixels) from the kernel center, $N$ is a normalization constant,
$R(\theta)$ is a 2D rotation matrix, and $\Sigma^{-1}$ is the inverse covariance (precision) matrix.
The covariance matrix is defined as
\begin{equation}
   \Sigma =
   \begin{bmatrix}
       \sigma_x^2 & 0 \\
       0 & \sigma_y^2
   \end{bmatrix},
\end{equation}
where $\sigma_i,\ i\in{x,y}$, denotes the spatial extent of the blur along the two image directions, and $R(\theta)$ determines the orientation of the blur kernel, thus controlling the coupling between horizontal and vertical smoothing. When  $\sigma_x = \sigma_y$ the Gaussian blur is isotropic and smooths the image equally in all directions. Otherwise, the blur is anisotropic, producing direction-dependent smoothing.
In our setup, $k_s^{(GB)}\sim [3,21],\ \text{and}\
\sigma^{(GB)}_x, \sigma^{(GB)}_y \sim [0.1,2.8]$.

\noindent
- The \textbf{Generalized Gaussian blur} kernel $\mathrm{k}^{(GG)}$ with size $k_s^{(GG)}$ can simulate a wide range of blur patterns. In the case of super-resolution, it has been shown that applying such kernels to degrade training images encourages sharper reconstructions on real-world settings \cite{Wang_2021_ICCV}. Formally,
\begin{equation}
   \mathrm{k}^{(GG)}(x,y) = \frac{1}{N} \exp \left(-\frac{1}{2} \left(C^T R^T(\theta) \Sigma^{-1}
   R(\theta) C\right)^\beta \right),\ C=[x,y]^T
\end{equation}
where $\beta$ is the shape parameter determining the kurtosis of a distribution, and $R(\theta), \Sigma,\ \text{and}\ N$ denote the rotation matrix, covariance matrix and normalization constant, respectively, as in the Gaussian blur. Taking $\beta=1$ gives a standard Gaussian distribution, where $\beta<1$ creates heavier tails, resulting in more uniform, box-like blurs. On the other hand, $\beta>1$ creates lighter tails, resulting in more peaked, focused blurs.
In our experiments $k_s^{(GG)} \sim [3,21],\quad \sigma^{(GG)}_x, \sigma^{(GG)}_y, \beta^{(GG)} \sim [0.5, 8]$, and $\theta \sim [0,2\pi]$. Finally, multiplicative noise ($\sigma_n \sim [0.9, 1.1]$) is added to the kernel to introduce more variation.

\noindent
- \textbf{Motion blur} is modeled as a convolution between the image and a line-shaped kernel that represents the path of movement during exposure \cite{levin_understanding_nodate, brooks_learning_2019}. A line-shaped kernel $\mathrm{k}^{(MB)}_{N,\theta,d}$ with controllable size $k$, angle $\theta^{(MB)}$, and direction range $d$ is used to emulate motion blur as:
\begin{equation}
\begin{aligned}
\mathrm{k}^{(MB)}_{N,\theta,d}&(x,y)
=
\frac{1}{k}
\sum_{n=0}^{k-1}
\boldsymbol{1}\!\left(
C
- t_n(d)\,u(\theta)
\right)\, \\
t_n(d)=t_{-}(d) + &\frac{n}{k-1}(t_{+}(d) -t_{-}(d)),\ n=0,\dots,k-1
\\
t_-(d)=-&\frac{k-1}{2}(1-d), \quad
t_+(d)= \ \frac{k-1}{2}(1+d), \\
\theta \sim \mathcal{U}(\theta_{\min},\theta_{\max}), 
\
&d \sim \mathcal{U}(d_{\min},d_{\max}) \subseteq [-1,1],
\
C=[x,y],\
u(\theta)=
\begin{bmatrix}
\cos\theta\\
\sin\theta
\end{bmatrix}.
\end{aligned}
\end{equation}
In our setting, $k\sim [3,21],\ \theta^{(MB)} \sim [0,2\pi],\ \text{and}\ d \sim [-1,1]$ 
Also, random shifts in kernel position are used, i.e. the kernel can be randomly offset from the center.

\noindent
- \textbf{Defocus blur} is synthesised by combining a disc-shaped kernel $\mathrm{k}^{(DB)}$ with radius $r$ and a Gaussian blur kernel $\mathrm{k}^{(DBG)}$, which together simulate the shape of a camera’s aperture.
Formally,
\begin{equation}
\mathrm{k}^{(DB)}(x,y) =
\begin{cases}
\frac{1}{\pi r^2}, & \text{if } x^2 + y^2 \le r^2,\\
0, & \text{otherwise,}
\end{cases}
\end{equation}
where $r$ denotes the blur radius proportional to the degree of defocus.
In our setup, we employ defocus blur using a circular kernel with $r\sim[3, 21]$. The Gaussian kernel size $k_s^{(DBG)}$ depends on the radius, i.e.\ 
$k_s^{(DBG)} = \mathbf{3}_{r\leq8} + \mathbf{5}_{r> 8}$, with $\sigma^{(DB)} \sim [0.1, 0.5]$.

\noindent
\emph{\textbf{Downscaling:}}
We use three different downsampling methods, namely nearest-neighbour, bilinear, and bicubic, with scaling factor $s \in\{2,4\}.$
Note that we do not explicitly use a blur kernel before downscaling to care for aliasing, unless it is randomly selected in the pipeline. However, we note that the averaging process of bilinear and bicubic interpolation inherently introduces mild blurring effects. Finally, in order to upscale an image to its original size, bicubic interpolation is used.\\[0.2em]
\emph{\textbf{Noise \& artifacts:}}
For the additive white Gaussian noise, we assume zero-mean Gaussian noise with standard-deviation per channel $\sigma_{GN} \sim [4\times 10^{-3}, 10^{-2}]$ for normalised inputs. 
%
Finally, the JPEG compression employed has a compression degree, which is controlled by a integer quality factor $q\in[0,100]$. Higher values of $q$ correspond to better visual quality and lower compression, and vice-versa \cite{zhang_designing_2021}. In our pipeline,  we used $q\sim[30, 95].$

\subsubsection{Learning-based degradations}
We employ the CUT \cite{park_contrastive_2020} GAN to learn a data-driven degradation model by translating high-quality images into degraded-quality ones. The training procedure follows the original implementation, with the addition of a re-ID loss term. The re-ID loss, $L_{\mathrm{re-id}}$, is computed using a frozen, pre-trained re-ID CNN that measures the embedding distance between the original and generated images, encouraging the generated degradations to preserve animal identity. We conducted a small hyperparameter search to determine an appropriate weight, $\lambda_{\mathrm{re-id}}$, for $L_{\mathrm{re-id}}$. The objective was to preserve identity while avoiding excessive constraints on the translation process. Based on this search, we set $\lambda_{\mathrm{re-id}} = 0.01$.

We then use the trained model to generate synthetic degraded images, which are included in the training set of the augmented re-ID model. For each training sample, a synthetic degraded counterpart is generated, effectively doubling the size of the training set. The CUT model is trained exclusively on the training split defined throughout this paper, ensuring that no information leakage occurs between training and testing.

\subsection{Data splitting details}
We provide detailed information about the dataset split which was used in our experiments. As mentioned in Section~3, we have three subsets, the training set, the set of individuals that are only part of the search database, and the query set.
Table ~\ref{fig:dataset_details} provides the number of images and the number of individuals for each of those categories and for each dataset. 

\begin{table}[t]
\resizebox{0.8\textwidth}{!}{
\begin{tabular}{@{}l
@{\hspace{12px}}c@{\,|\,}c
@{\hspace{12px}}c@{\,|\,}c
@{\hspace{12px}}c@{\,|\,}c@{}}
\toprule
& \multicolumn{2}{c}{\textbf{Training}}
& \multicolumn{2}{c}{\textbf{Database Only}}
& \multicolumn{2}{c}{\textbf{Query}} \\
\cmidrule(lr){2-3}\cmidrule(lr){4-5}\cmidrule(lr){6-7}
& \textbf{Images} & \textbf{IDs}
& \textbf{Images} & \textbf{IDs}
& \textbf{Images} & \textbf{IDs} \\
\midrule
ATRW (tigers)        & 3,472 & 152  & 870  & 30  & 1,073 & 166 \\
BelugaID             & 5,393 & 661  & 1,349 & 127 & 1,817 & 719 \\
CTai (chimpanzees)   & 2,857 & 56   & 872  & 15  & 933  & 65  \\
ELPephants           & 1,309 & 226  & 337  & 48  & 432  & 248 \\
Giraffes             & 878  & 145  & 223  & 33  & 267  & 161 \\
GiraffeZebraID       & 4,371 & 1,693 & 1,095 & 358 & 1432 & 1029 \\
HumpbackWhaleID      & 7,897 & 1,886 & 1,976 & 379 & 2930 & 2054 \\
HyenaID2022          & 1,998 & 214  & 505  & 42  & 626  & 235 \\
IPanda50             & 4,401 & 41   & 1,100 & 9   & 1,373 & 46  \\
LeopardID2022        & 4,247 & 369  & 1,153 & 61  & 1,406 & 358 \\
MacaqueFaces         & 4,021 & 27   & 1,005 & 7   & 1,254 & 30  \\
NyalaData            & 1,204 & 200  & 320  & 37  & 418  & 216 \\
OpenCows2020         & 3,016 & 37   & 773  & 9   & 947  & 42  \\
SealID               & 1,323 & 46   & 338  & 11  & 419  & 52  \\
SeaTurtleID2022      & 5,579 & 362  & 1,399 & 76  & 1,751 & 397 \\
StripeSpotter        & 520  & 36   & 135  & 9   & 165  & 40  \\
WhaleSharkID         & 4,899 & 461  & 1,229 & 82  & 1,565 & 473 \\
ZindiTurtleRecall    & 7,897 & 1,886 & 1,976 & 379 & 2930 & 2054 \\
\midrule
\textbf{TOTAL}       & \textbf{67,078} & \textbf{10,741}
                     & \textbf{17,137} & \textbf{2,208}
                     & \textbf{22,354} & \textbf{8,996} \\
\bottomrule
\vspace{0.5em}
\end{tabular}

}
\caption{The number of images and individuals (images | ids) in each subset. The first column refers to the training set of the model. The second refers to individuals in the image retrieval database but not seen during training, and the third to images in the query set.}
\label{fig:dataset_details}
\end{table}

\subsection{Model implementation details}
\subsubsection{Transformer-Based re-ID model}
The model used for most experiments is the large Swin variant transformer (Swin-L), which uses a patch size of $4\times4$, a window size of $12$, and an input resolution of $384\times384$ pixels. The backbone produces a $1536$-dimensional feature embedding for each input image.

The model is trained using the CurricularFace loss \cite{huang_curricularface_2020}, an adaptive margin-based softmax loss for discriminative feature learning. The loss projects feature vectors onto a unit hypersphere and enforces an angular margin between different classes. Unlike other popular margin-based losses, such as ArcFace \cite{deng_arcface_2022} and SphereFace \cite{liu_sphereface_2017}, which apply a fixed angular margin and treat all samples equally, CurricularFace dynamically adjusts the decision boundaries according to both the training stage and the difficulty of individual samples.

The CurricularFace loss function, $\mathcal{L}$, is defined as
\begin{equation}    
\begin{aligned}
   \mathcal{L}_(\theta_{y_i}, \theta_j;t) = -\log & \frac{ e^ {s \cos (\theta_{y_i} + m )} }{ e^ {s \cos (\theta_{y_i} + m )} + \sum_{j=1,\ j \neq y_i} ^n  e^{sN\left(t^{(k)},\ \cos\theta_j \right)} }, \\
   N(t, \cos\theta_j) & = 
   \begin{cases}
       \cos\theta_j,& T(\cos\theta_{y_i}) - \cos\theta_j \geq 0, \\
       \cos\theta_j(t+\cos\theta_j),& T(\cos\theta_{y_i}) - \cos\theta_j < 0,
   \end{cases}\\
   T(\cos\theta_{y_i}) &= \cos(\theta_{y_i}+m).
\end{aligned}
\end{equation}
Here $s$ is a scaling factor that controls the magnitude of logits, and $m$ is the additive angular margin that enforces inter-class separation. For each sample $i$, $\theta_{y_i}$ denotes the angle between the extracted feature vector and the target class $y_{i}$, while $\theta_j$ represents the angles between the same vector and any other class. The function $N(t, \cos\theta_j)$ adaptively adjusts the logits of the negative classes according to training stage and difficulty. Easy negatives remain unchanged, while hard negatives are scaled up as $t$ increases. The parameter $t$ is computed via an exponential moving average of the positive cosine similarities within each batch $k$. We use $m=0.5$ and $s=64$.

To determine suitable values for the learning rate, momentum, and the loss function parameters 
$s$ and $m$, we performed a small-scale hyperparameter search using the Optuna framework \cite{optuna_2019}.

To train this model, we employ stochastic gradient descent with a learning rate of $3\times10^{-3}$ and momentum $0.9$. A cosine annealing scheduler is employed to gradually reduce the learning rate during training. Models are trained with a batch size of $16$ and $4$ gradient accumulation steps, resulting in an effective batch size of 32. Training is terminated when the training loss and evaluation metrics flatten out, 150 epochs for the baseline and simple model, and 200 for the diverse ones.

\subsubsection{CNN-based re-ID model}
This model follows the implementation described in \cite{otarashvili_multispecies_2024}, using an EfficientNetV2-M backbone \cite{pmlr-v139-tan21a}. Unlike the original implementation, the input image resolution is $384\times384$, which the network maps to $2048$-dimensional feature embeddings. The model is trained using the sub-center ArcFace loss \cite{vedaldi_sub-center_2020} with $k=3$ sub-centers per class. Apart from the increased input resolution, the training procedure is identical to that described in the original paper.


\subsection{Full version of Figure 5 (main text)}
In Figure \ref{fig:results_arti_degs} we provide a version of Figure 5 (main text) where we have also included the diverse model.
\begin{figure}[t]  
\begin{center}
\begin{minipage}[c]{0.75\textwidth}
    \centering
    \vspace{-1.5em}
    \includegraphics[width=0.99\textwidth,trim={0 0 1.0cm 0},clip]{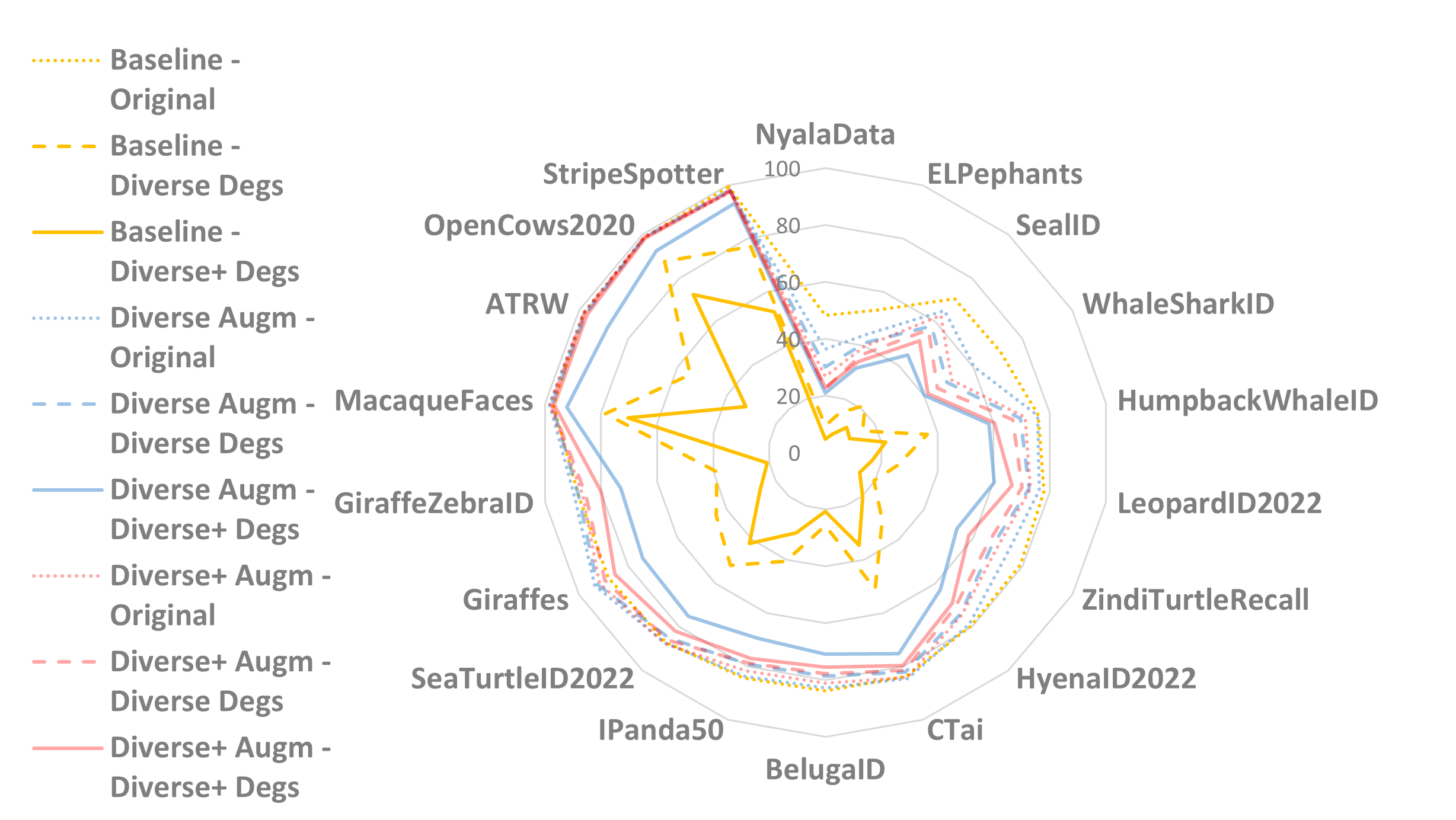}
\end{minipage}   
\end{center}
    \caption{Full version of Figure 5 in the main text, including here the diverse model as well. Rank-1 accuracies for the same experiments reported separately for each of the 18 datasets, where we observe that the magnitude of the adverse effects of image degradations on re-ID performance is species-specific.
    }
    \label{fig:results_arti_degs}
\end{figure}

\subsection{Additional results on re-ID performance on real-world degraded images}
\subsubsection{Model-based augmentation for transformer-based re-ID}
In Section~4.2 we discussed the re-ID performance on real-world data regarding query images of clarity  1 and 4, as well as for all query data. In addition, here Figure~\ref{fig:TurtleHeads_clarity_2,3} presents the performance for clarity levels 2 and 3. In the time-unaware setting, there is comparable Rank-$k$ performance of all models for both clarity levels. In the time-aware setting, the augmented models show a slight improvement at Rank-1 and Rank-5. The diverse model achieves the best overall performance at Rank-1, although at Rank-20 and clarity level 2 the baseline slightly outperforms the augmented models.

\begin{figure*}[htb]
 
\begin{minipage}[t]{0.45\textwidth}  
   \centering
   \textbf{Time-unaware, all individuals}
\end{minipage}
\begin{minipage}[t]{0.45\textwidth}  
   \centering
      \textbf{Time-aware,   not seen in training}
\end{minipage}
\vspace{0.2em}

\begin{minipage}[t]{0.02\textwidth} 
\vspace{-2.5cm}
\rotatebox{90}{Clarity 2}\\
\vspace{1.5cm}
\rotatebox{90}{Clarity 3}\\
\end{minipage}
\begin{minipage}[t]{0.45\textwidth}   
\centering
    \includegraphics[width=0.99\linewidth]{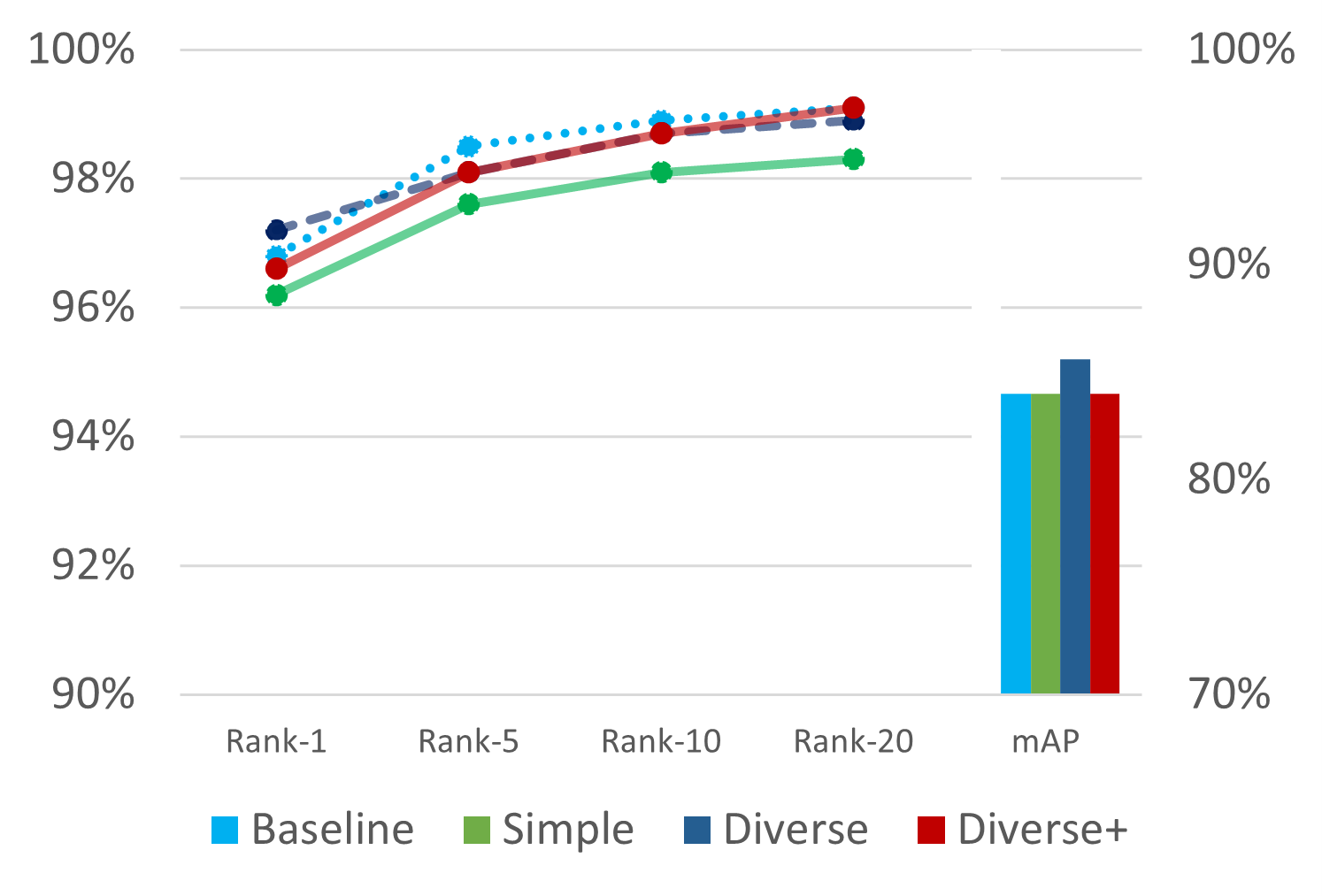}\\
    \includegraphics[width=0.99\linewidth]{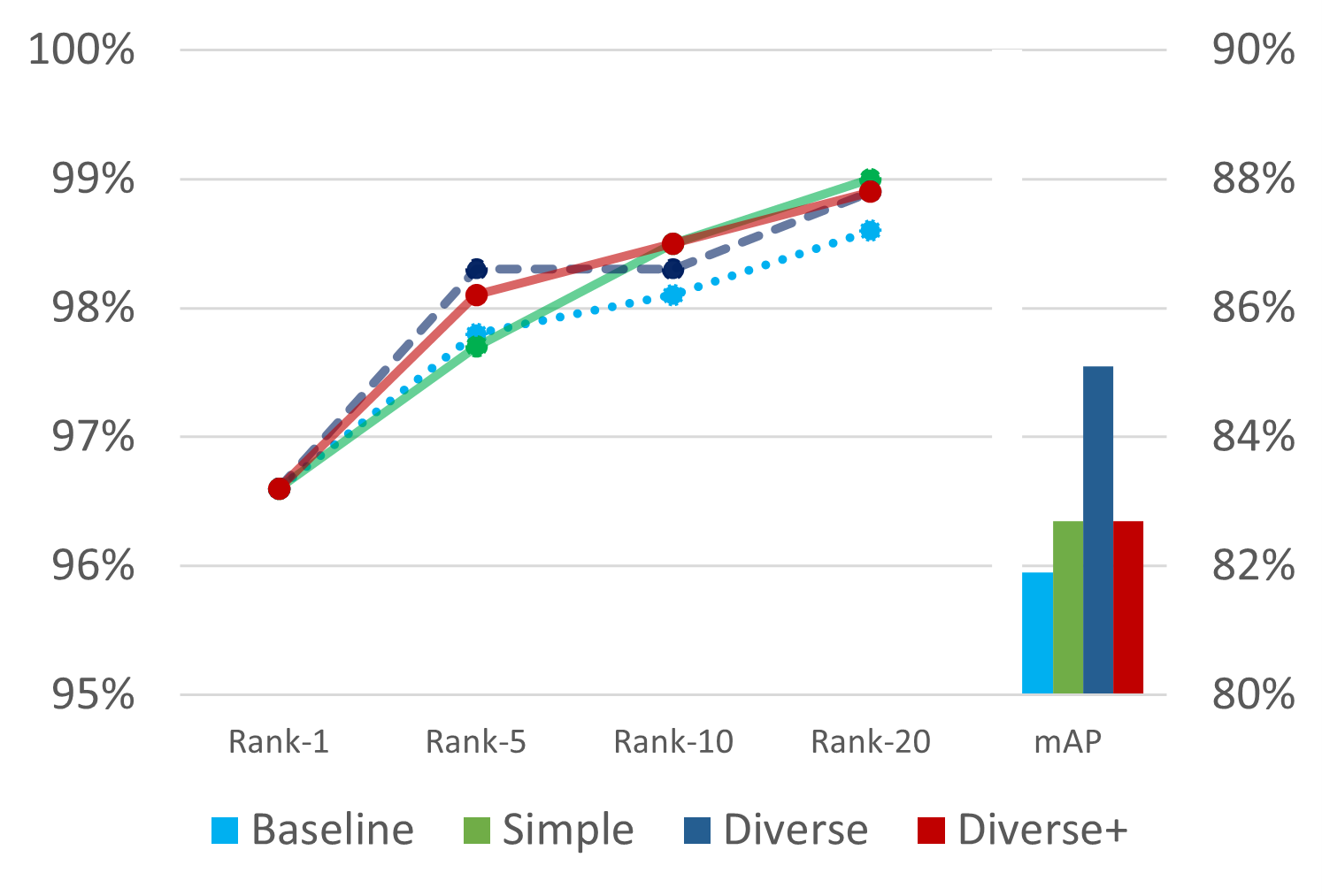}\\
\end{minipage}
\begin{minipage}[t]{0.45\textwidth}  
\centering
     \includegraphics[width=0.99\linewidth]{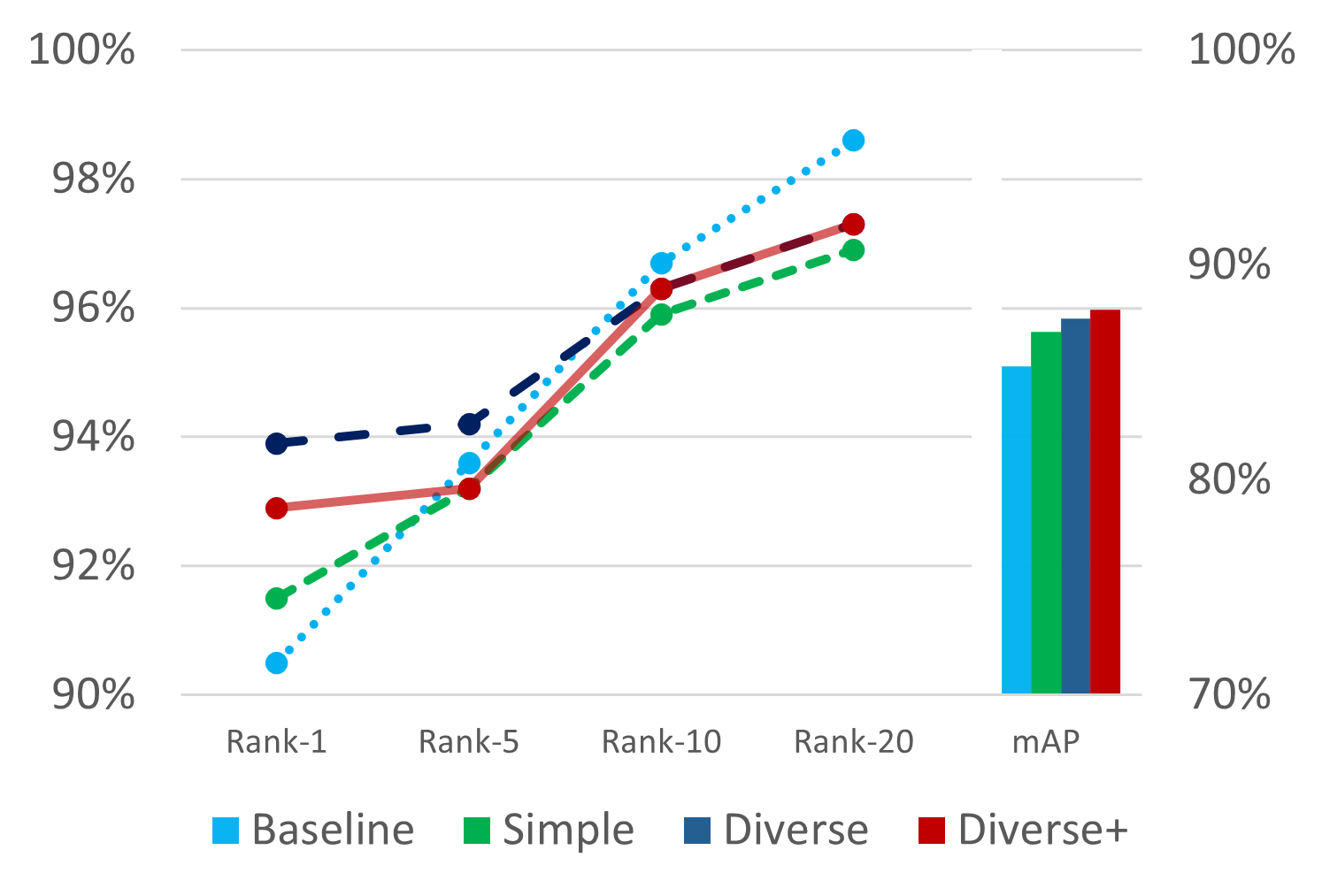}\\
    \includegraphics[width=0.99\linewidth]{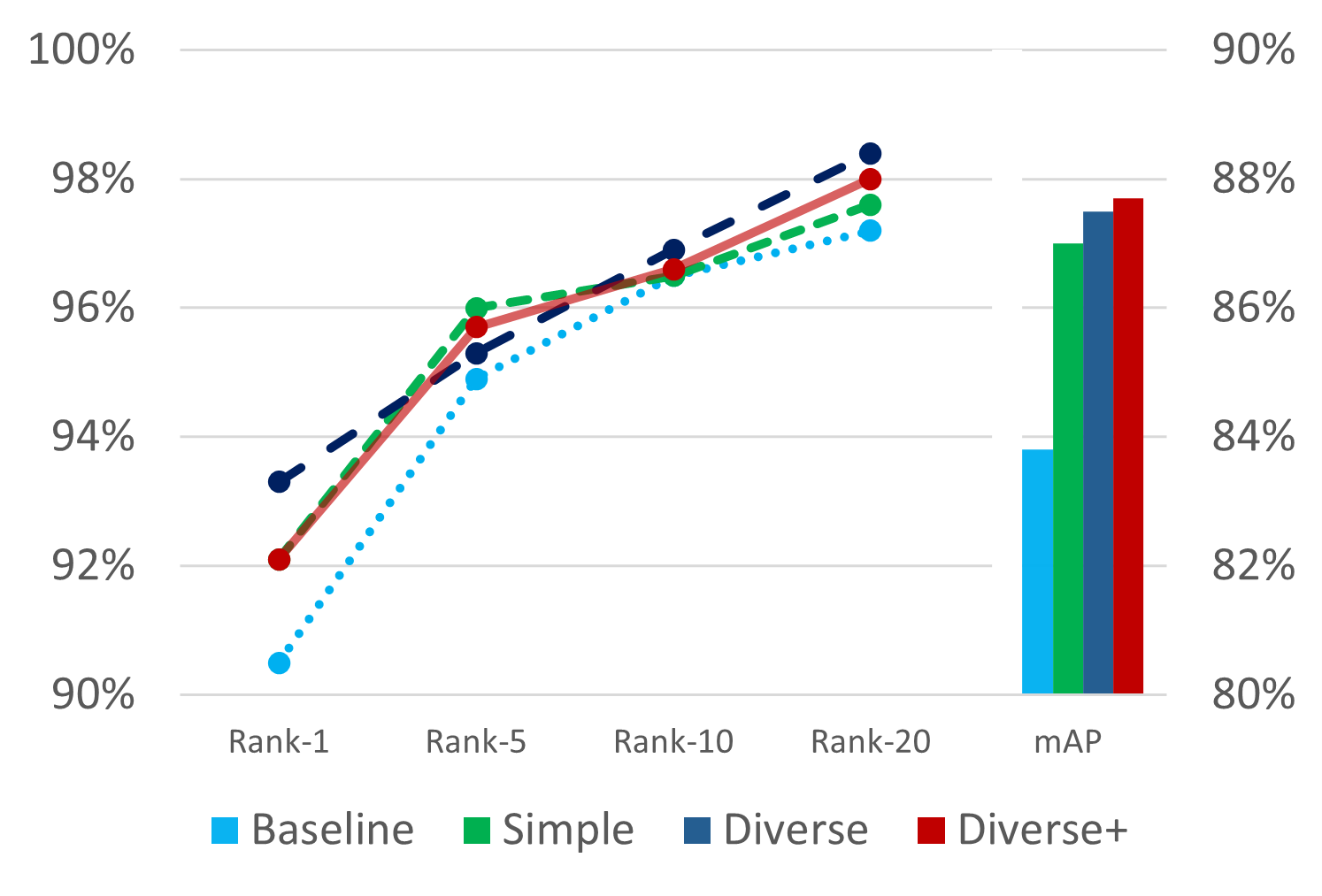}\\
\end{minipage}
    \caption{
    Top: Re-ID performance  for query images of clarity 2, 3 of SeaTurtleID2022. 
    At the left column  the query set is restricted to individuals not seen during training and whose images are split into database and query in a time-aware fashion.
    }
    \label{fig:TurtleHeads_clarity_2,3}
\end{figure*}

\begin{figure*}[t]
\centering
\includegraphics[width=0.33\textwidth]{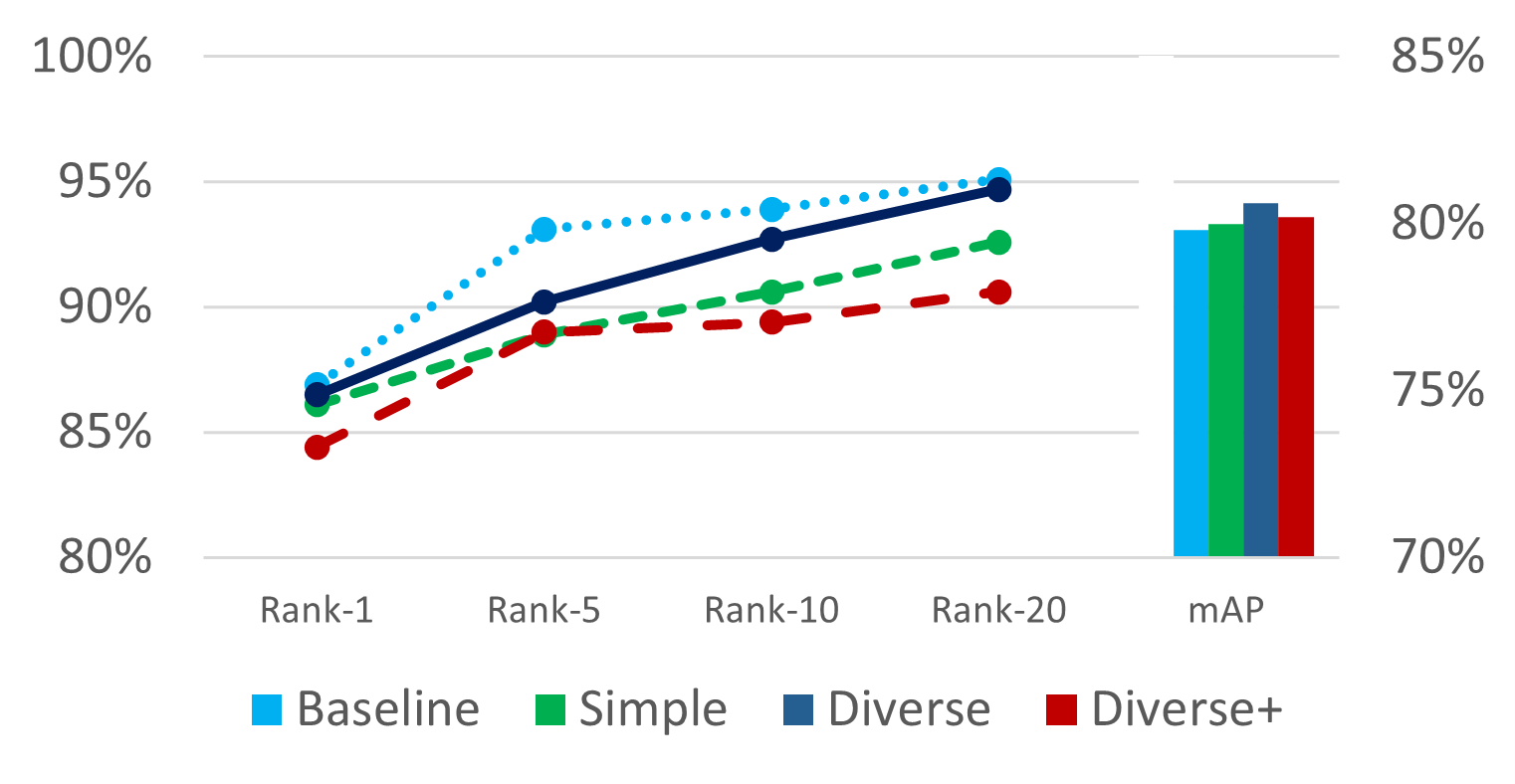}\hfill
\includegraphics[width=0.33\textwidth]{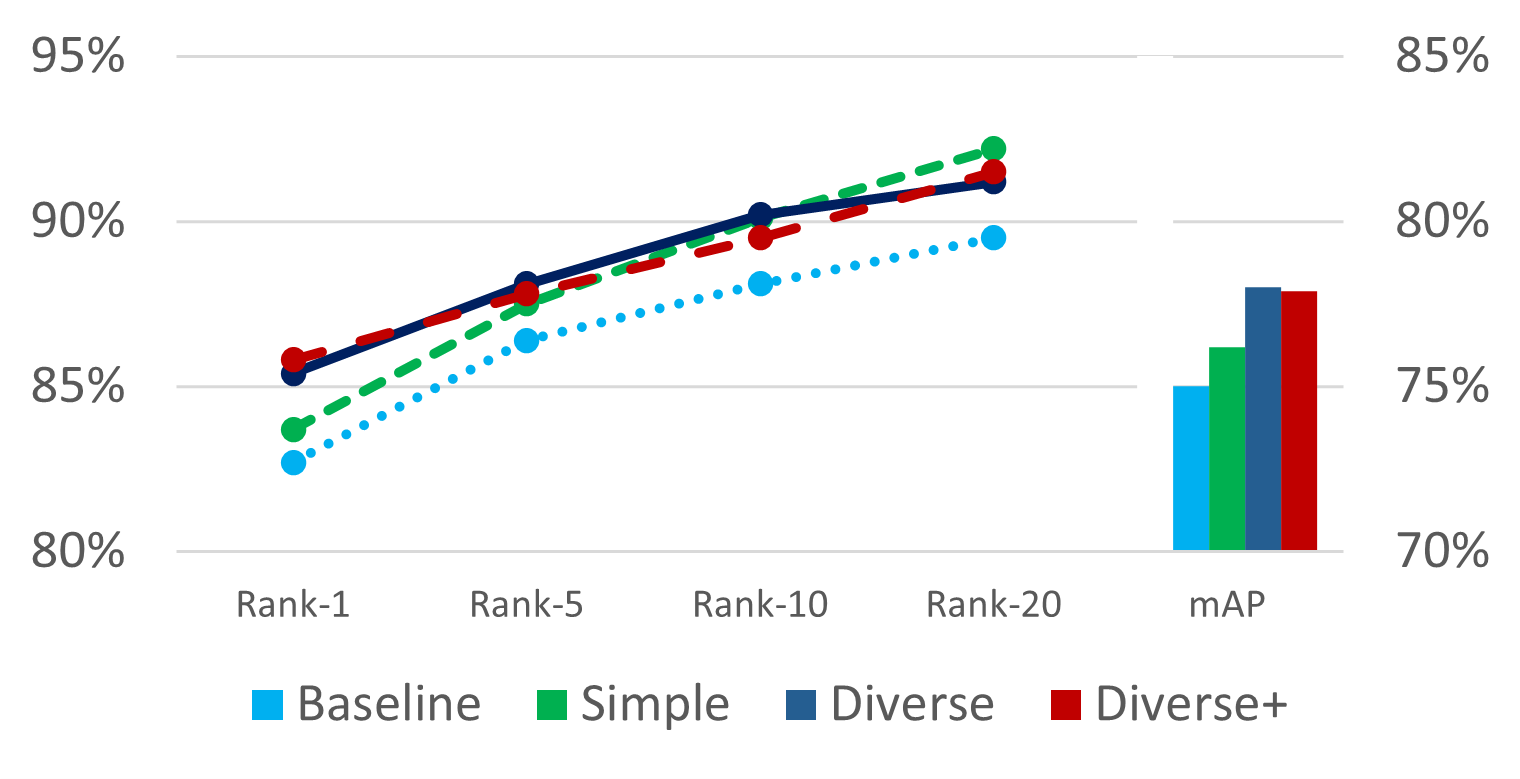}\hfill
\includegraphics[width=0.33\textwidth]{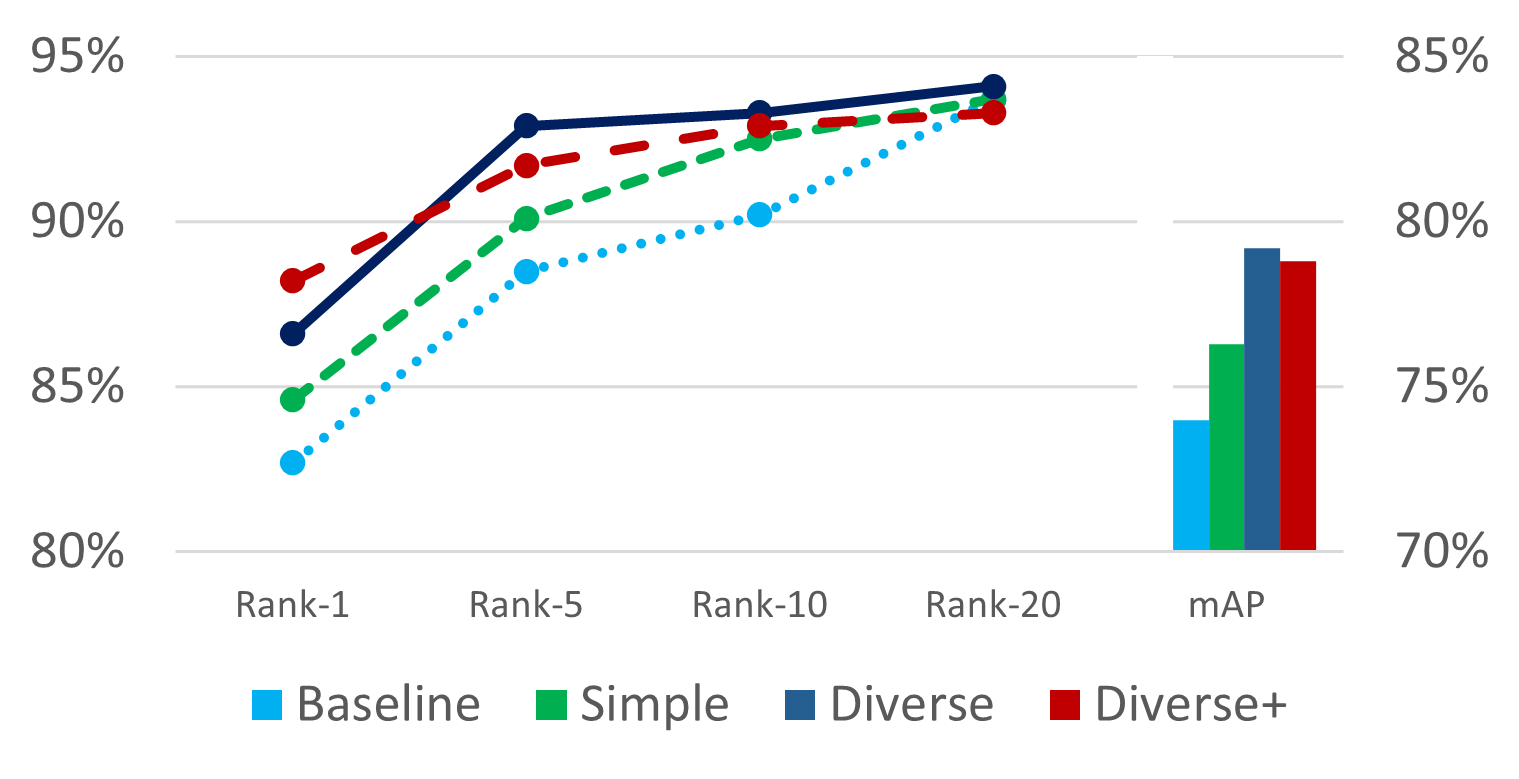}

\vspace{2mm}

\includegraphics[width=0.48\textwidth]{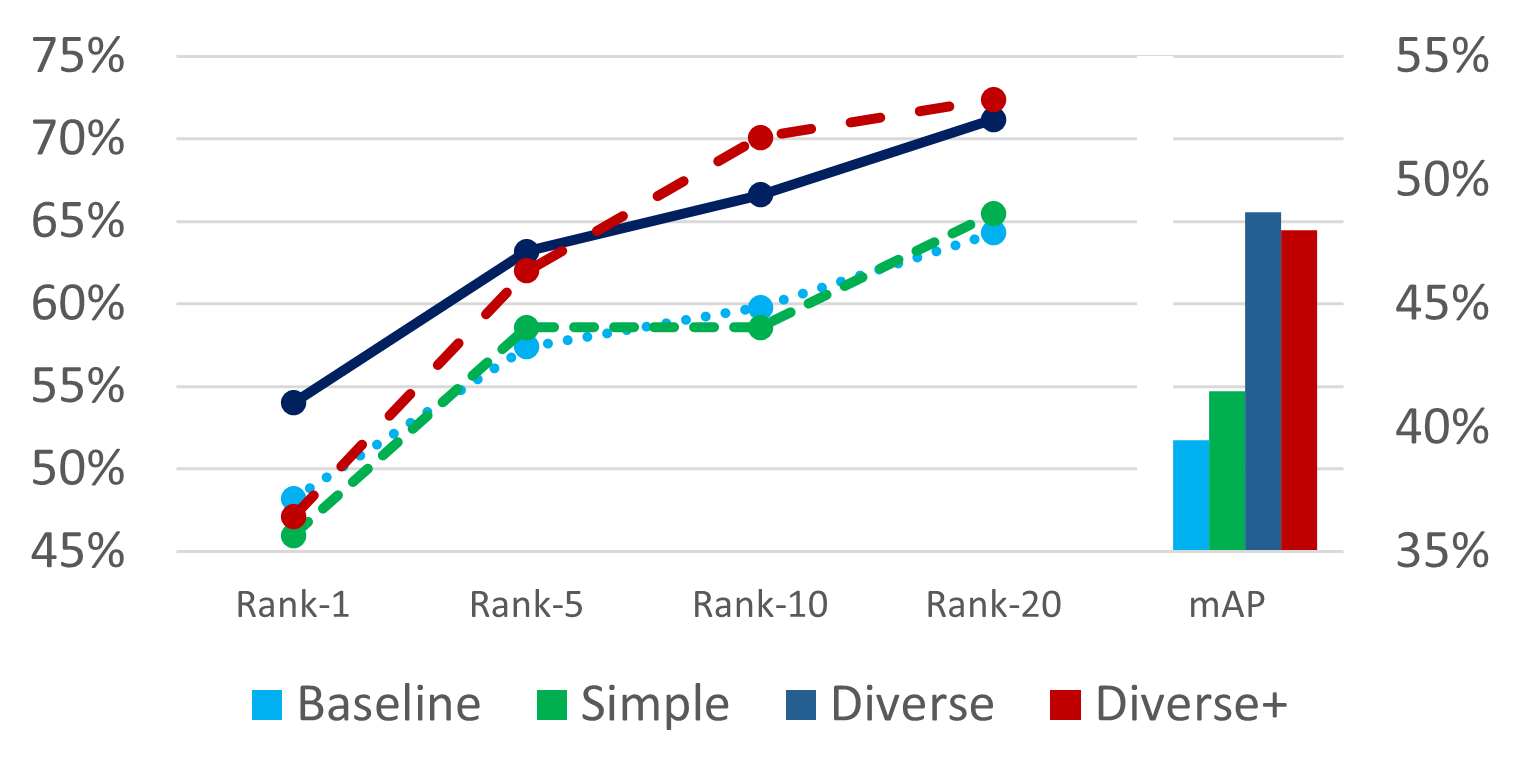}\hfill
\includegraphics[width=0.48\textwidth]{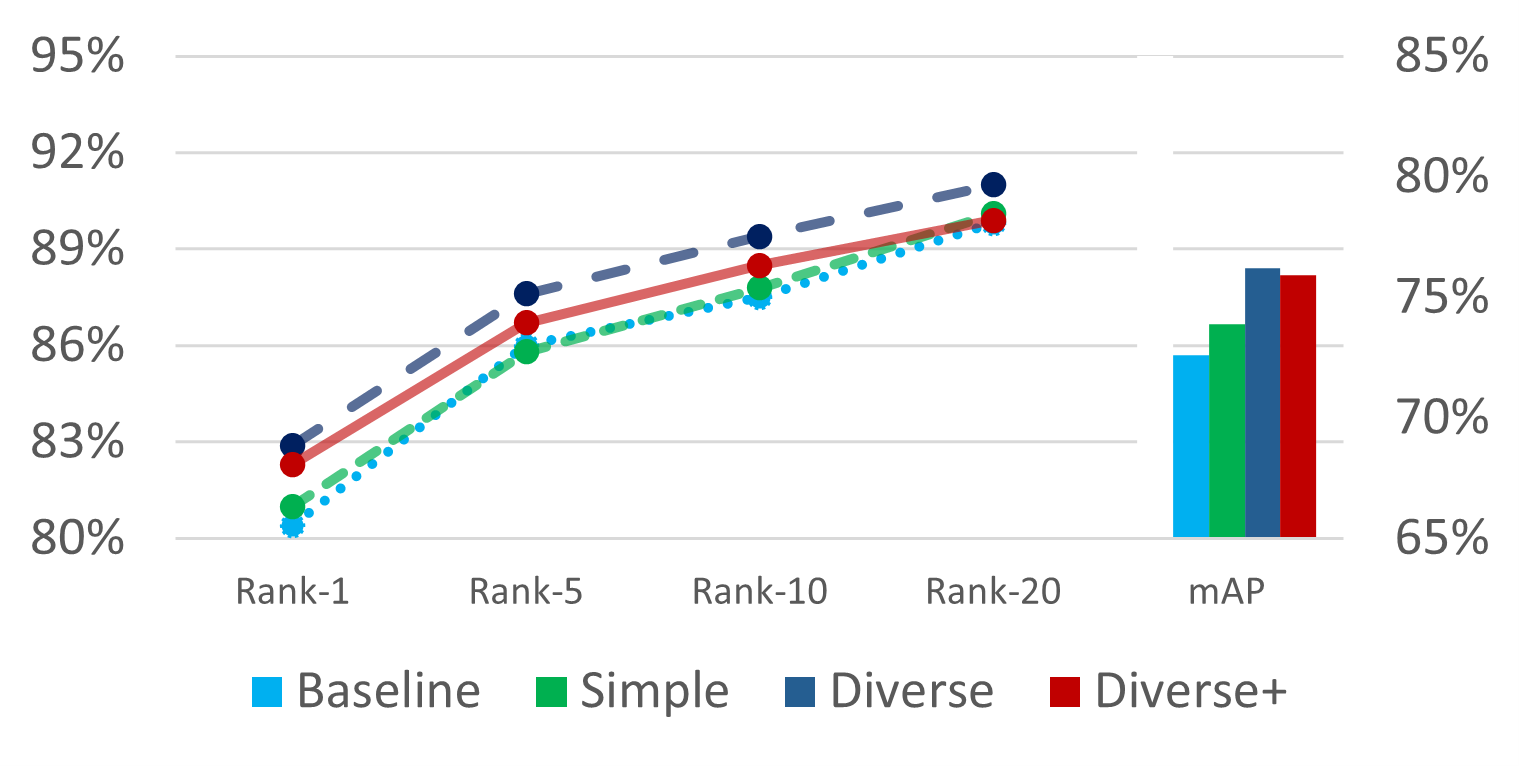}

\caption{
Top: Re-ID performance for query images at clarity levels 1–4 and their average on SeaTurtleID2022.
The query is restricted to individuals not see during training, whose images have been re-split into a database and query set in a \emph{time-aware} fashion. The difference to Figure \ref{fig:TurtleHeads_clarity_2,3} is that here  also the individuals from the training set have been included to the database.
}
\label{fig:TurtleHeads_clarity_time_aware_full}
\end{figure*}

\noindent
In Figure \ref{fig:TurtleHeads_clarity_time_aware_full}, we also report results  for the setting in which the query is restricted to individuals not seen during training and is re-split in a time-aware fashion, (left column of Figure  \ref{fig:TurtleHeads_clarity_2,3}), but now additionally including the images of individuals from the training set to the image retrieval database. Since these additional individuals do not correspond to any identity in the query set, they solely increase the difficulty of the retrieval task. We observe similar results: Improved performance of the augmented models over the baseline, overall and in all clarity sets but 1 with particularly increased performance in clarities 3 and 4.

\subsubsection{Learning-based augmentation for CNN-Based re-ID}
\begin{figure*}[!htb]
\begin{minipage}[t]{0.47\textwidth}  
   \centering
   \textbf{Time-unaware, all individuals}
\end{minipage}
\begin{minipage}[t]{0.47\textwidth}  
   \centering
      \textbf{Time-aware,   not seen in training}
\end{minipage}
\vspace{0.2em}

\begin{minipage}[t]{0.02\textwidth} 
\vspace{-2.5cm}
\rotatebox{90}{Clarity 1}\\
\vspace{1.5cm}
\rotatebox{90}{Clarity 4}\\
\vspace{1.1cm}
\rotatebox{90}{Full query set}
\end{minipage}
\begin{minipage}[t]{0.45\textwidth}   
\centering
    \includegraphics[width=0.99\linewidth]{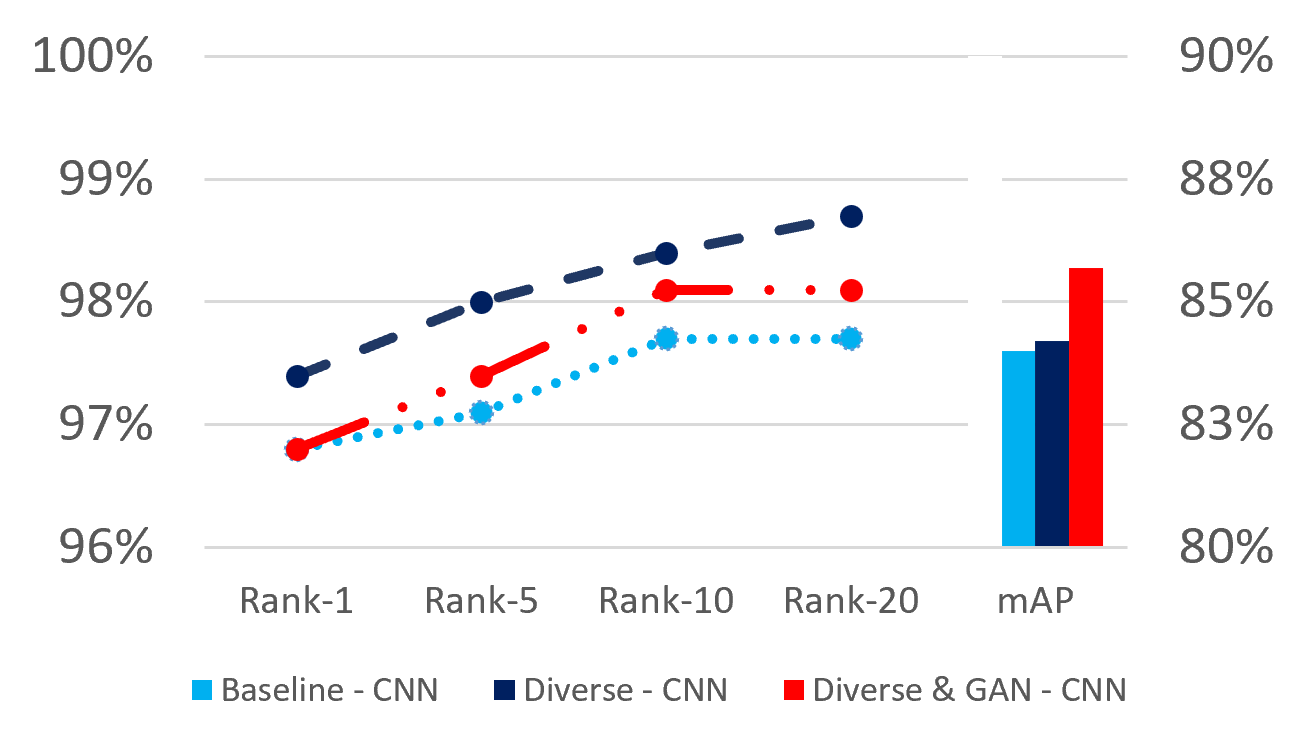}\\
    \includegraphics[width=0.99\linewidth]{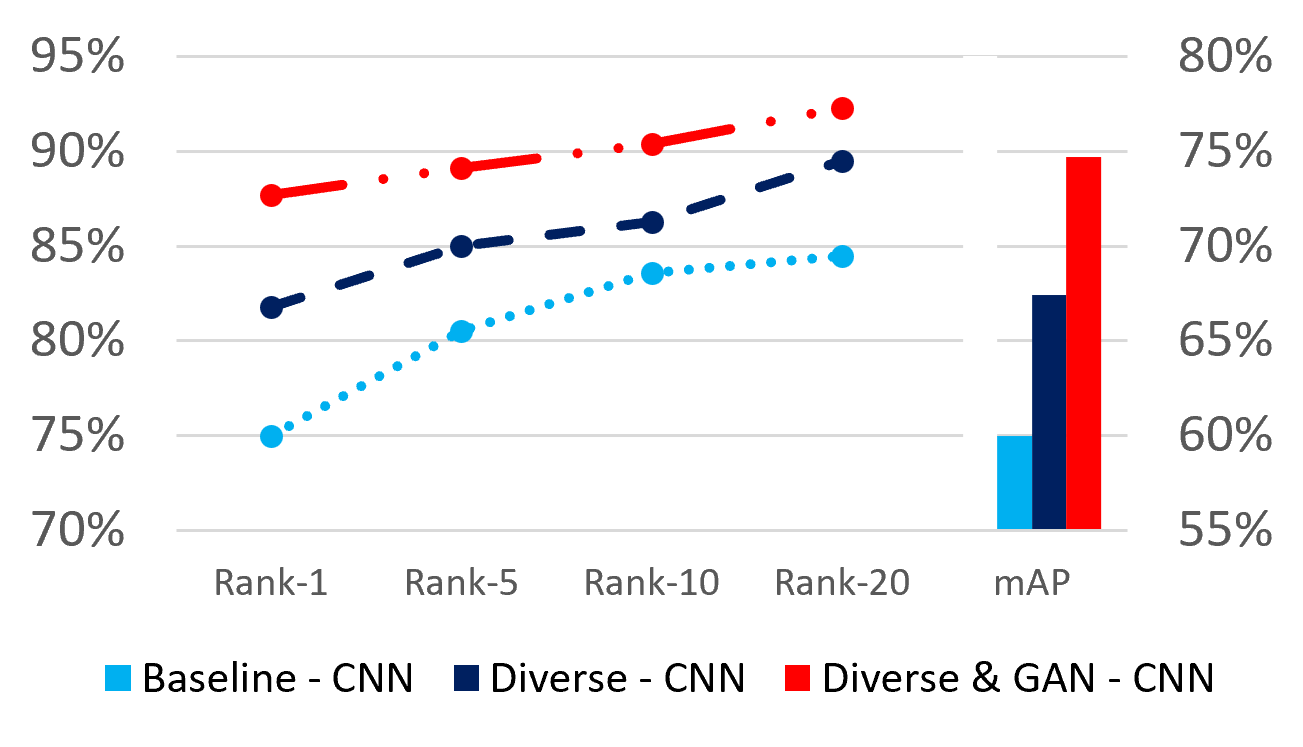}\\
    \includegraphics[width=0.99\linewidth]{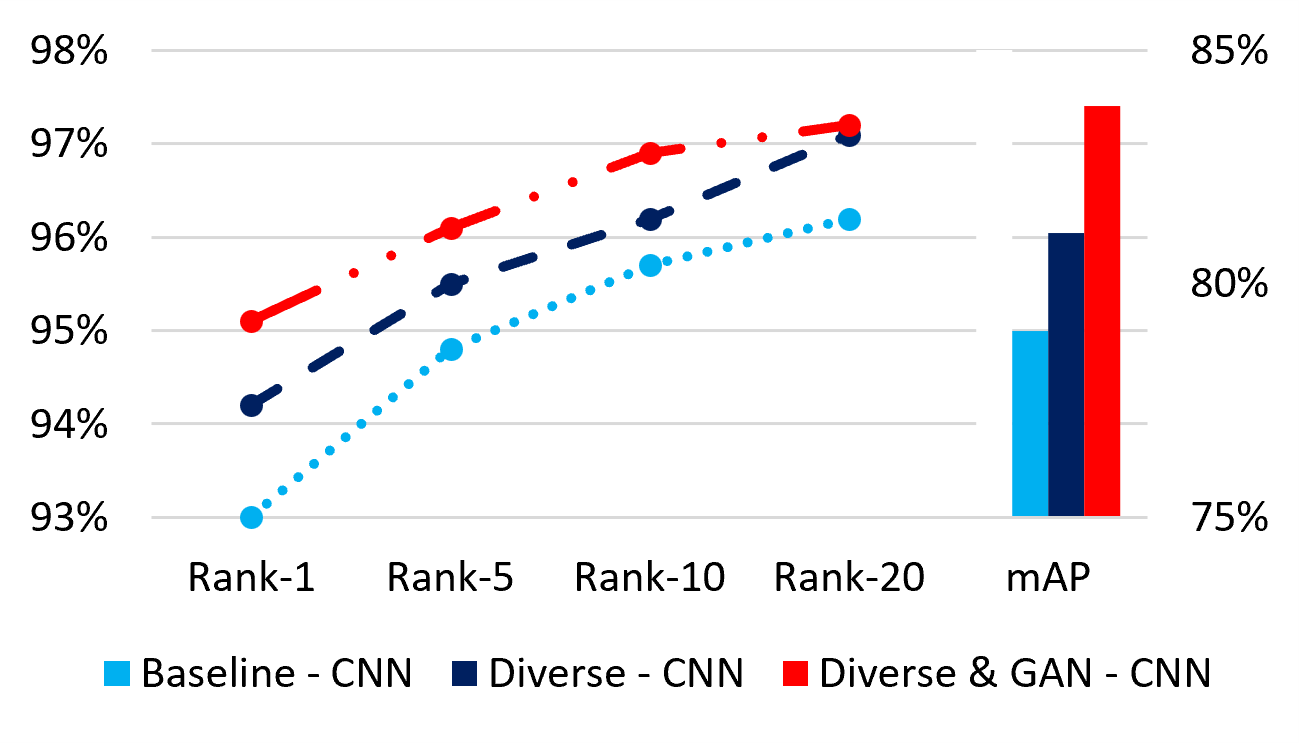}
\end{minipage}
\begin{minipage}[t]{0.45\textwidth}  
\centering
     \includegraphics[width=0.99\linewidth]{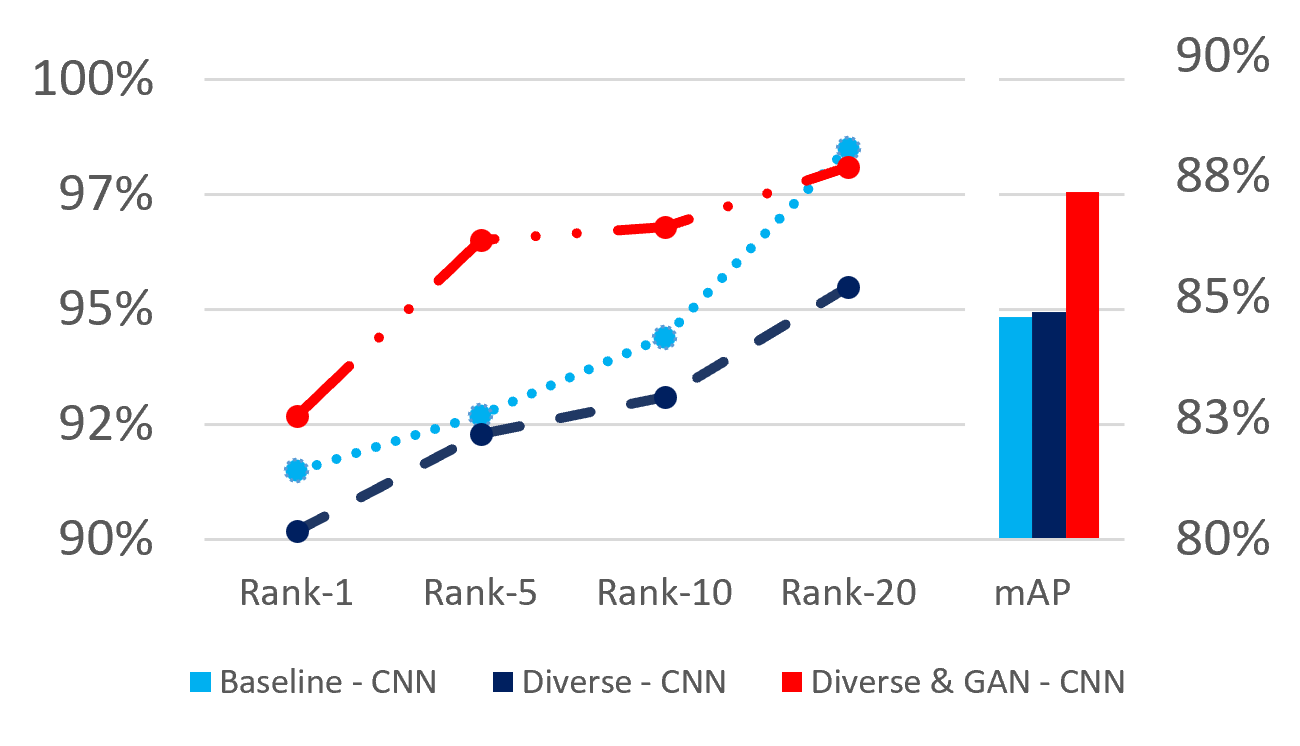}\\
    \includegraphics[width=0.99\linewidth]{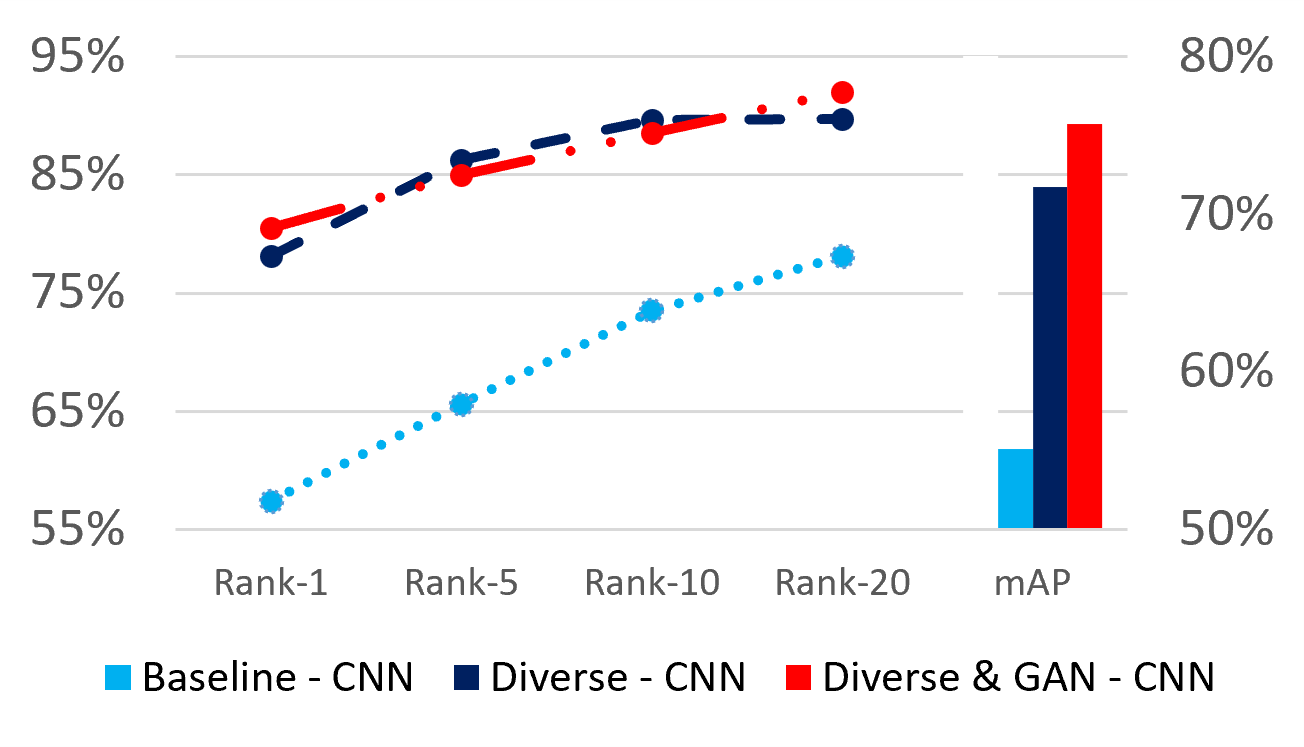}\\
    \includegraphics[width=0.99\linewidth]{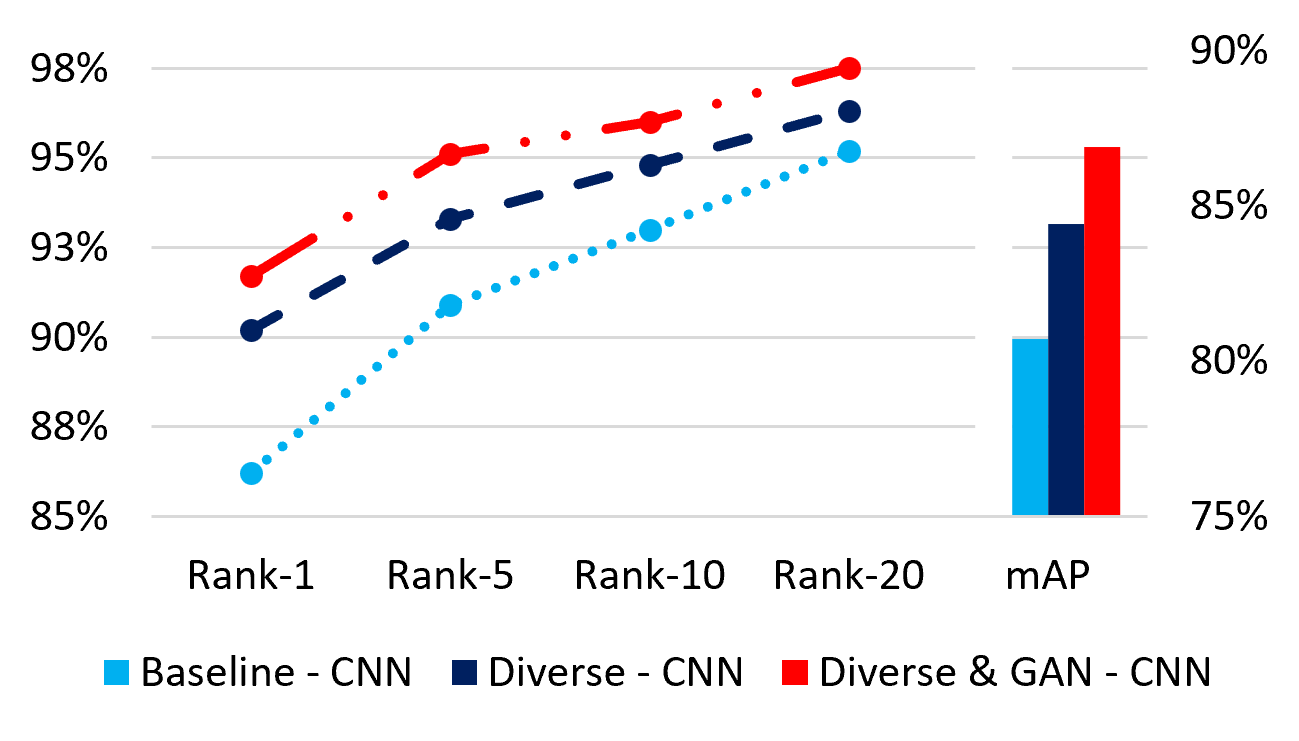}
\end{minipage}
    \caption{Re-ID performance with CNN backbone for query images of clarity 1, 4 and the full query set of SeaTurtleID2022.}
    \label{fig:TurtleHeads_clarity_cnn}
\end{figure*}

In Figure~\ref{fig:TurtleHeads_clarity_cnn}, we report results in the same format as Figure~6 of the main text, but using a CNN backbone. We compare a baseline model, a model augmented using the model-based diverse pipeline only, and the one further augmented model trained with learning-based synthetic data. The latter is the one  reported in the main text.

\subsection{Comparison with MegaDescriptor on unseen datasets}
\begin{figure}[ht]
\begin{minipage}[t]{0.99\textwidth} 
\centering
\includegraphics[width=0.7\textwidth,trim={0 0.5cm 1.3cm 0.3cm},clip]{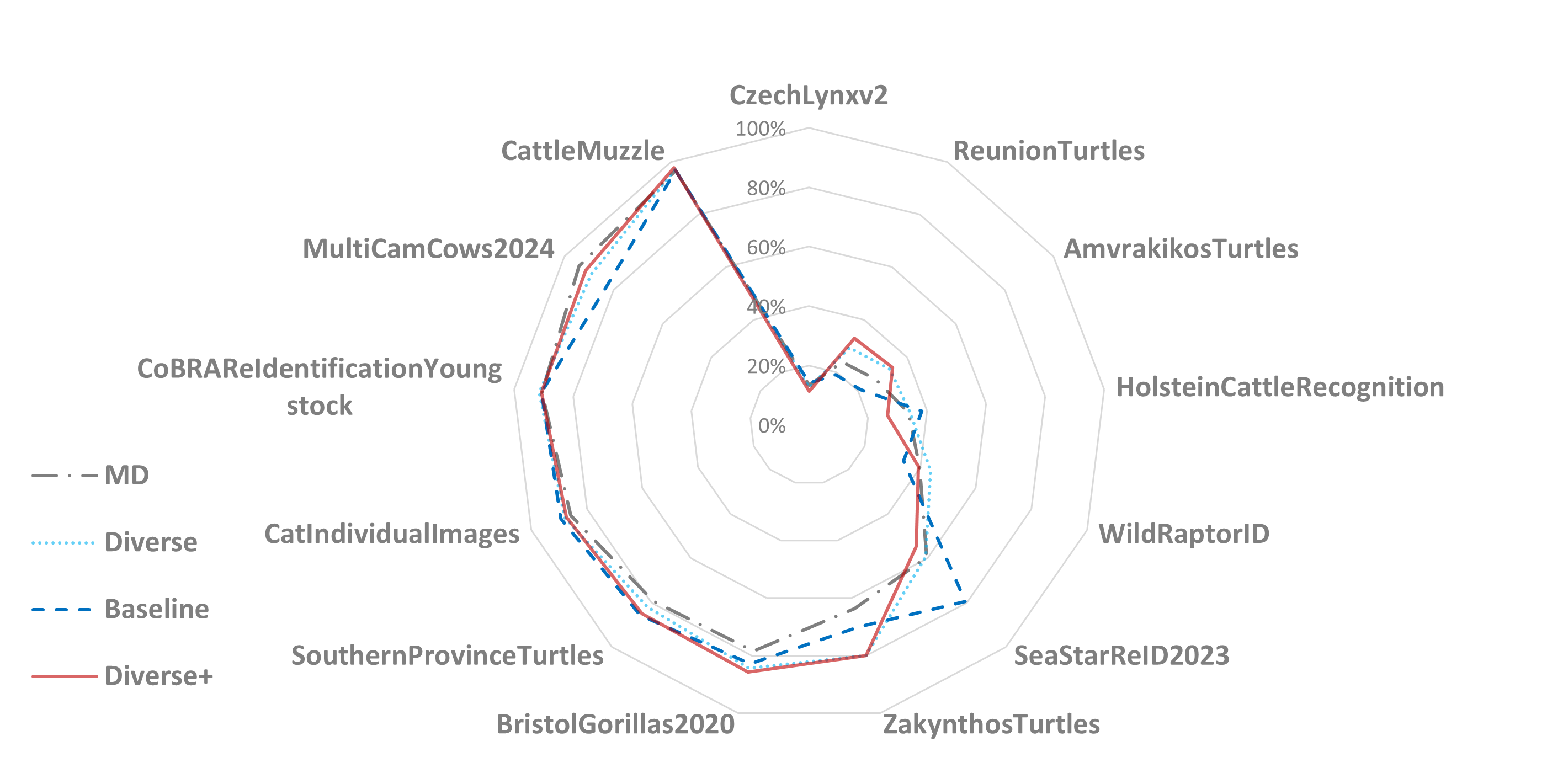}
\end{minipage}
    \caption{Rank-$1$ accuracies for baseline, augmented models, and MegaDescriptor (MD), for datasets that were not seen during training by any model. In most datasets, the augmented models improve upon or are on par with the baseline and MegaDescriptor.}
    \label{fig:comparisons}
\end{figure}

We avoided comparing with state-of-the-art deep feature-based multi-species re-ID models such as MegaDescriptor and MiewID \cite{cermak_wildlifedatasets_2024, otarashvili_multispecies_2024}, since the SeaTurtleID2022 dataset has been seen during training for these models under a different split and thus there is a danger of train-to-test data leak. We also stress again that here our focus is to show the benefit of degradation-based augmented training for low quality images, rather than introducing a new foundation model. Because of these reasons and also due to its training set not being publicly available, we avoided comparisons with MiewID \cite{otarashvili_multispecies_2024}. However, in order to check that our augmented models are on par with state-of-the-art performance, we perform some last re-ID experiments including MegaDescriptor, using 13 datasets from the WildlifeDatasets library that none of the models has seen during training. We refer to the library for corresponding references. We performed a 70/30 search database-query set split which was done in the time-aware fashion whenever timestamps were available.
In Figure \ref{fig:comparisons}, we provide Rank-$1$ scores for the baseline, diverse, diverse$^{+}$ models and MegaDescriptor. In most datasets, the augmented models improve upon or are on par with the baseline as well as MegaDescriptor, which has even been trained with more datasets. 

\end{document}